\documentclass[hidelinks]{article}

\usepackage{PRIMEarxiv}

\usepackage[utf8]{inputenc} 
\usepackage[T1]{fontenc}    
\usepackage{hyperref}       
\usepackage{url}            
\usepackage{booktabs}       
\usepackage{amsfonts}       
\usepackage{nicefrac}       
\usepackage{microtype}      
\usepackage{lipsum}
\usepackage{fancyhdr}       
\usepackage{graphicx}       
\graphicspath{{media/}}     

\usepackage{amssymb}
\usepackage{url}
\usepackage{hyperref}
\usepackage{multirow}
\usepackage[small]{caption}

\usepackage{enumerate}
\usepackage[shortlabels]{enumitem}
\usepackage[authoryear]{natbib}

\usepackage{algorithm}
\usepackage[noend]{algpseudocode}

\algdef{SE}[DOWHILE]{Do}{doWhile}{\algorithmicdo}[1]{\algorithmicwhile\ #1}%
\newcommand{\A}{{\cal A}}

\newcommand{\M}{{\cal M}}

\newcommand{\zug}[1]{\langle #1  \rangle}
\newcommand{\stam}[1]{}
\newcommand{\set}[1]{\{ #1  \}}

\newcommand{\Rat}{\mathbb{Q}}

\newtheorem{remark}{Remark}
\newtheorem{example}{Example}
\newtheorem{theorem}{Theorem}

\title{ASQ-IT: Interactive Explanations for Reinforcement-Learning Agents}

\author{
  Yotam Amitai, Ofra Amir\\
  Faculty of Data \& Decision Science\\
  Technion - I.I.T, Haifa\\
  \texttt{yotama@campus.technion.ac.il}\\
  \texttt{oamir@technion.ac.il} \\
   \And
  Guy Avni \\
  Faculty of Computer Science \\
  University of Haifa, Haifa \\
  \texttt{gavni@cs.haifa.ac.il} \\
}

\begin{document}
\maketitle

\begin{abstract}
As reinforcement learning methods increasingly amass accomplishments, the need for comprehending their solutions becomes more crucial.  
Most explainable reinforcement learning (XRL) methods generate a static explanation depicting their developers' intuition of what should be explained and how. In contrast, literature from the social sciences proposes that meaningful explanations are structured as a dialog between the explainer and the explainee, suggesting a more active role for the user and her communication with the agent.
In this paper, we present ASQ-IT -- an interactive explanation system that presents video clips of the agent acting in its environment based on queries given by the user that describe temporal properties of behaviors of interest. Our approach is based on formal methods: queries in ASQ-IT's user interface map to a fragment of Linear Temporal Logic over finite traces (LTLf), which we developed, and our algorithm for query processing is based on automata theory. 
User studies show that end-users can understand and formulate queries in ASQ-IT and that using ASQ-IT assists users in identifying faulty agent behaviors.
\end{abstract}


\section{Introduction}
\label{sec:intro}
Reinforcement Learning (RL) has shown impressive success in recent years; e.g., mastering Go, achieving human-level performance in Atari games, and more recently, in the development of advanced AI systems like ChatGPT by utilizing human feedback ~\cite{silver2016mastering,mnih2015human,christiano2017deep}.
However, current training techniques are complex and rely on implicit goals and implicit feature representations, and thus largely produce black-box agents. In order for such trained agents to be successfully deployed, in particular in safety-critical domains such as healthcare or transportation, it is crucial for them to be trustworthy; namely, both developers and users need to understand, predict and assess agents' behavior. This need has led to an abundance of ``explainable RL'' (XRL) methods~\cite{dazeley2021explainable} designed to elucidate black-box agents. 

Existing XRL methods predominantly offer static explanations, detailing the agent's behavior or decision process through means such as saliency maps~\cite{greydanus2017visualizing}, causal explanations~\cite{madumal2020explainable}, or simplified policy representations like decision trees~\cite{liu2019toward}. These approaches, however, lack interactivity, denying users the opportunity to engage with the information or ask specific questions. Moreover, the generated explanations address researcher-formulated questions rather than user-generated queries, reflecting the researchers' perspective of relevant issues rather than the users'.

Drawing on social sciences literature~\cite{miller2018explanation}, we focus on developing interactive XRL methods that foster dialogue between the system (explainer) and user (explainee) through repeated user queries. Recent trends highlight interactive explanations as key to system intelligibility and user engagement~\cite{abdul2018trends}. Interaction and exploration are also shown to mitigate over-reliance on AI~\cite{buccinca2021trust}. These insights emphasize the importance of not just enhancing AI performance but also encouraging user engagement with explanations and systems through interaction and exploration.

We follow an ``explanation by demonstration'' approach~\cite{amitaisurvey}: our system provides the user with clips of the agent interacting with its environment. It has been shown that using such explanation systems is helpful for users' assessment of agent behavior~\cite{amir18highlights,sequeira2020interestingness,huber2020local,amitai2021don,septon2023integrating,amitai2023explaining}.
The key challenge in this approach is that a user's attention span is very limited, thus the system must carefully select the clips that convey the most information about the agent. 
Previously, various techniques have been studied to select which traces are shown such as state importance~\cite{amir18highlights,huang2018establishing}, ability to generalize a policy~\cite{huang2017enabling,lage2018evaluation} or agent disagreements~\cite{amitai2021don}. All of these heuristics are static and do not allow user input. Moreover, the heuristic is chosen by the system designer, thus the user has no control over it.


In this work, we develop ``ASQ-IT'' - \textbf{A}gent \textbf{S}ystem \textbf{Q}ueries \textbf{I}nteractive explana\textbf{T}ions, that aims to assist users to comprehend an agent in a global manner, by generating clips of the agent interacting with its environment. 
The user controls which clips will be presented by formulating queries that specify properties of clips of interest. The interaction with the explanation system resembles a dialogue: the user enters a query, and receives clips that match it; the user can then refine her query, and the process continues.

One of the key challenges in developing an interactive explanation system is the interaction with human users (especially laypeople). 
An explanation system's query interface must strike the right balance between expressivity and usability. That is, the system should allow users to express a wide variety of queries while maintaining a simple query mechanism that users will understand. Moreover, since we are concerned with describing behavior in sequential decision-making settings, the query interface must allow users to reason about temporal behaviors.

We address these challenges by taking a formal approach. 
An established logic to reason about temporal properties is Linear Temporal Logic (LTL) \cite{Pnu77}. A primary application for LTL is verification ({\em model checking}~\cite{CE81,handbookMC}): the specification of a system (i.e., the correct behaviors) is expressed using an LTL formula, and the goal is to formally prove that a model of the system satisfies the specification. LTL reasons on infinite traces, and we use a more recent logic called LTLf~\cite{LTLf-patterns}, which reasons about finite traces. ASQ-IT allows users to express the clips of the agent that they are interested in observing via an LTLf formula. 
This raises a challenge: ASQ-IT is intended for laypeople with no background in logic. To make the system accessible, we develop a simple subclass of LTLf and a user interface that maps user queries to LTLf formulas in our fragment. Our backend consists of a library of clips that are collected offline. Given a user query (as an LTLf formula), we apply an automata-based algorithm to search the library of clips, and present to the user clips that matches their query. 

To evaluate ASQ-IT, we conducted two user studies. The first study assessed laypeople's ability to utilize and understand queries using the ASQ-IT interface. The second study explored the use of ASQ-IT by participants with some AI knowledge in an agent debugging task. We observe improved performance compared to a static policy summary baseline. More importantly, we qualitatively observed that participants interacting with ASQ-IT were more engaged in the task and were able to generate, validate, and refute their own hypotheses regarding the agent's behavior.

The paper makes the following contributions:
\begin{itemize}
    \item We propose a novel interactive explanation method for Reinforcement Learning (RL) agents, which involves retrieving traces of agent behavior that correspond to a specified temporal logic query.
    \item We introduce ASQ-IT, an interactive system designed to enable users to easily formulate queries, which are then converted into logical formulas. The system subsequently displays the relevant extracted clips.
    \item Our research demonstrates the usability of ASQ-IT, highlighting its accessibility even to individuals with no expert knowledge in the field.
    \item We provide evidence of the superiority of ASQ-IT's interactive approach over traditional static policy summary methods in the context of debugging tasks.
\end{itemize}

The paper is organized as follows. Section~\ref{sec:examples} presents running examples of the ASQ-IT system. Section~\ref{sec:related_work} discusses relevant related work and further highlights the distinction ASQ-IT. Section~\ref{sec:ASQIT_implement} details the design and technical aspects of ASQ-IT. This is followed by Section~\ref{sec:evaluation}, which covers the empirical evaluation of the system through user studies. Finally, Section~\ref{sec:summary_future_work} provides a summary of the paper along with possible future directions.

\section{ASQ-IT: Running examples}
\label{sec:examples}
Before delving into the technical details of ASQ-IT, we introduce two running examples that will be used to illustrate the ASQ-IT system, both of which were implemented and used in our user studies as described in Section~\ref{sec:evaluation}.


\subsection{Domains}
We first describe the domains ASQ-IT was implemented for.

\begin{figure}[ht]
	\centering
    \includegraphics[width=1\columnwidth]{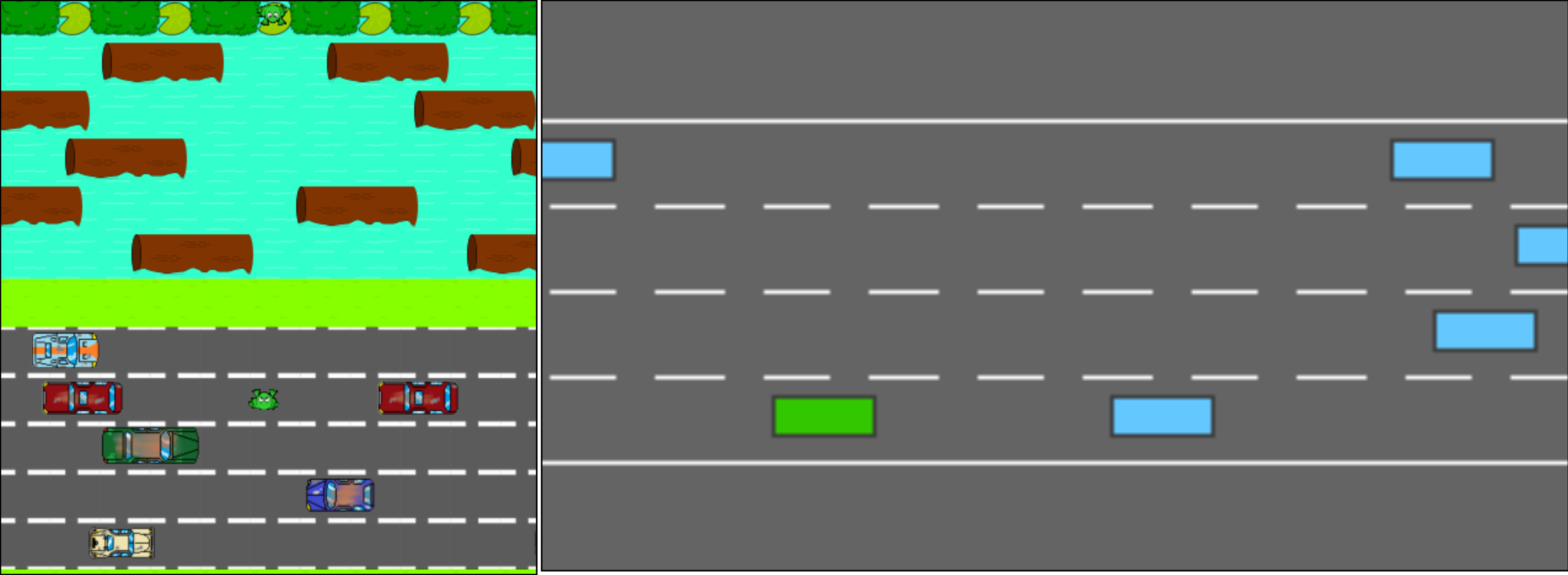}\\
	\caption{{\textbf{Left:} Frogger domain} - agent is a green frog that seeks to reach the lilypads at the top of the screen while avoiding cars on the road and jumping between logs on the river. \textbf{Right:} Highway domain - agent is a green rectangular vehicle driving on a multi-lane highway while interacting with other vehicles and avoiding collision.}
	\label{fig:domains}
\end{figure}

\paragraph{\textbf{Highway Domain\protect\footnote{Github repository: https://github.com/Farama-Foundation/HighwayEnv} (Fig.~\ref{fig:domains}-Right)}}
The domain consists of a multiple-lane highway in which the agent controls a car depicted as a green rectangle. Other uncontrollable cars are depicted as blue rectangles. Cars can accelerate, decelerate, and change lanes. We train agents for various goals by using different reward functions in training; for example, a combination of not crashing, driving fast, driving in the right lane, etc. The absence of a fixed objective and the ongoing dynamics of the environment add layers of complexity, making it an ideal testbed for advanced Explainable Reinforcement Learning (XRL) research~\cite{zhou2020smarts,feng2021intelligent,li2022metadrive}. Its strategic selection for user studies lies in its realistic yet comprehensible nature, allowing participants to grasp the scenario with minimal instruction.

\paragraph{\textbf{Frogger Domain\protect\footnote{Github repository: https://github.com/pedrodbs/frogger} (Fig.~\ref{fig:domains}-Left)}} 
Frogger is a well-known Atari game. The objective of the game is to guide a frog from the bottom of the screen to an empty lilypad at the top of the screen. The agent controls the frog. In each step it chooses one of four actions: up, down, left, or right, causing the frog to hop in the respective direction. The frog must cross two areas with two types of obstacles. First, it must cross a road with moving cars while avoiding being run over. Second, it must cross a river by jumping on passing logs while avoiding falling into the water. 

\subsection{ASQ-IT usage: Example implementations}
We present two \textit{possible} implementations of ASQ-IT, particularly those utilized in our user studies, for the specified domains and demonstrate their application. We emphasize that these are example manifestations of the system and that both the interface and predicate selection are customizable and can be modified for various audiences, from experts requiring more expressive options to laypeople benefiting from simplicity.

\begin{figure}[ht]
	\centering
    \frame{\includegraphics[width=1\columnwidth]{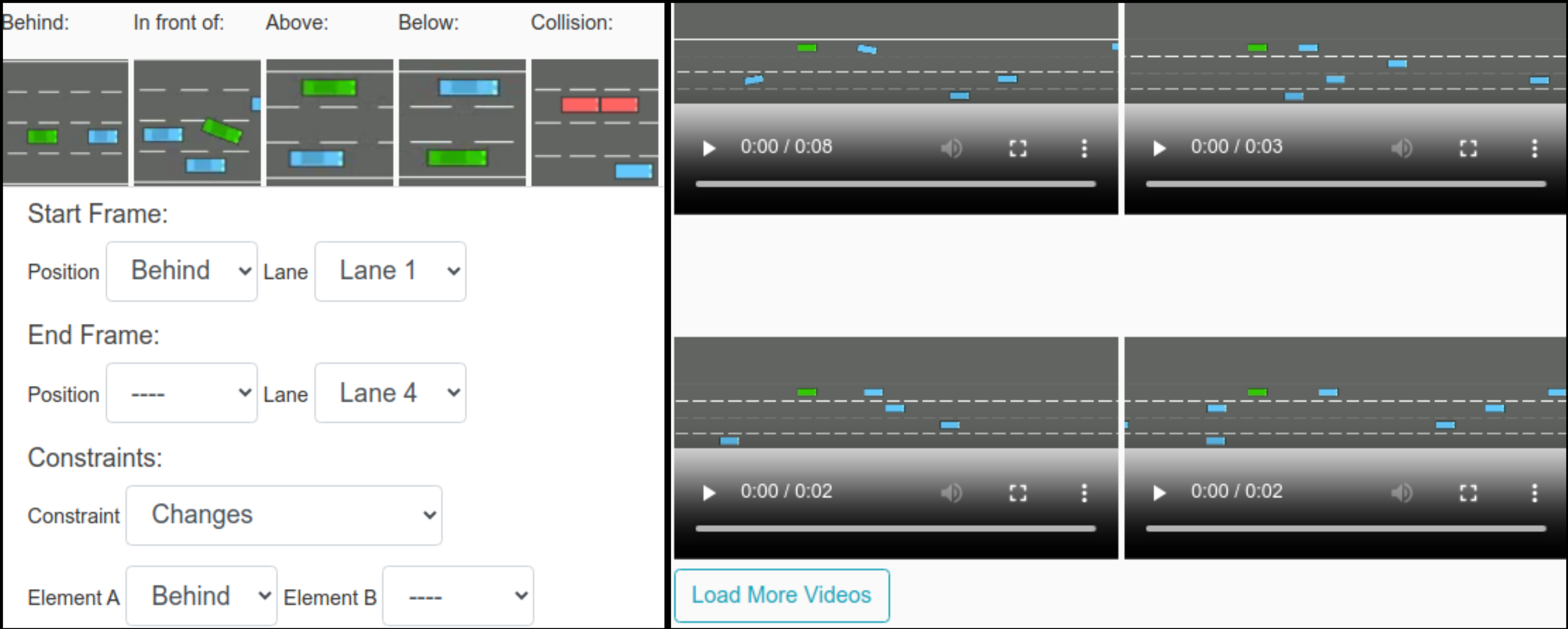}}\\
	\caption{Example ASQ-IT implementation for 
 the Highway domain. \textbf{Left:} Query interface; \textbf{Right:} Output generated videos. \textbf{Query interpretation:} ``Agent starts at lane 1 behind some car, and finishes at lane 4 while somewhere along the way, the agent is no longer behind a car.''}
	\label{fig: highway interface}
\end{figure}

\paragraph{\textbf{ASQ-IT Interface}}
The user's main interaction point with ASQ-IT is through the \emph{Query Interface} (see Fig.~\ref{fig: highway interface}-Left). 
The interface dictates the structure of a query. It allows users to constrain both the start and end frame of the clip that the system will output, as well as the clip as a whole. The user enters constraints in drop-downs that contain pre-determined predicates chosen by a domain expert, as we elaborate later. 

\begin{example}{\bf Highway Domain (Fig. \ref{fig: highway interface}).} 
\normalfont
Consider a user exploring the behavior of an autonomous agent in a driving simulation. In the first query, the user wants to understand how the agent behaves when changing lanes. They specify the agent's start state in ``Lane 1'' and the end state in ``Lane 4''. The system responds with clips showing the green car moving from the top lane to the bottom lane, illustrating lane-changing behavior. 
After reviewing these clips, the user becomes curious about the agent's behavior under more complex conditions. In the second query, they add a constraint: the agent must navigate behind another car for the entire duration of the clip while still moving from Lane 1 to Lane 4. This query is a direct development from the first, now focusing on the agent's ability to change lanes while maintaining a specific position relative to another car. The system then provides new clips, showcasing how the agent dynamically adjusts its lane-changing strategy when following another vehicle. 
Through this interactive process, the user gains deeper insights into the agent's decision-making and adaptability in varying traffic conditions.\footnote{Example of output video in \href{https://osf.io/hj3cu/?view_only=9ce523fd224741e3a4767003b146e2b5}{Supplementary Materials}}


\end{example}

\begin{figure}[ht]
	\centering
    \frame{\includegraphics[width=1\columnwidth]{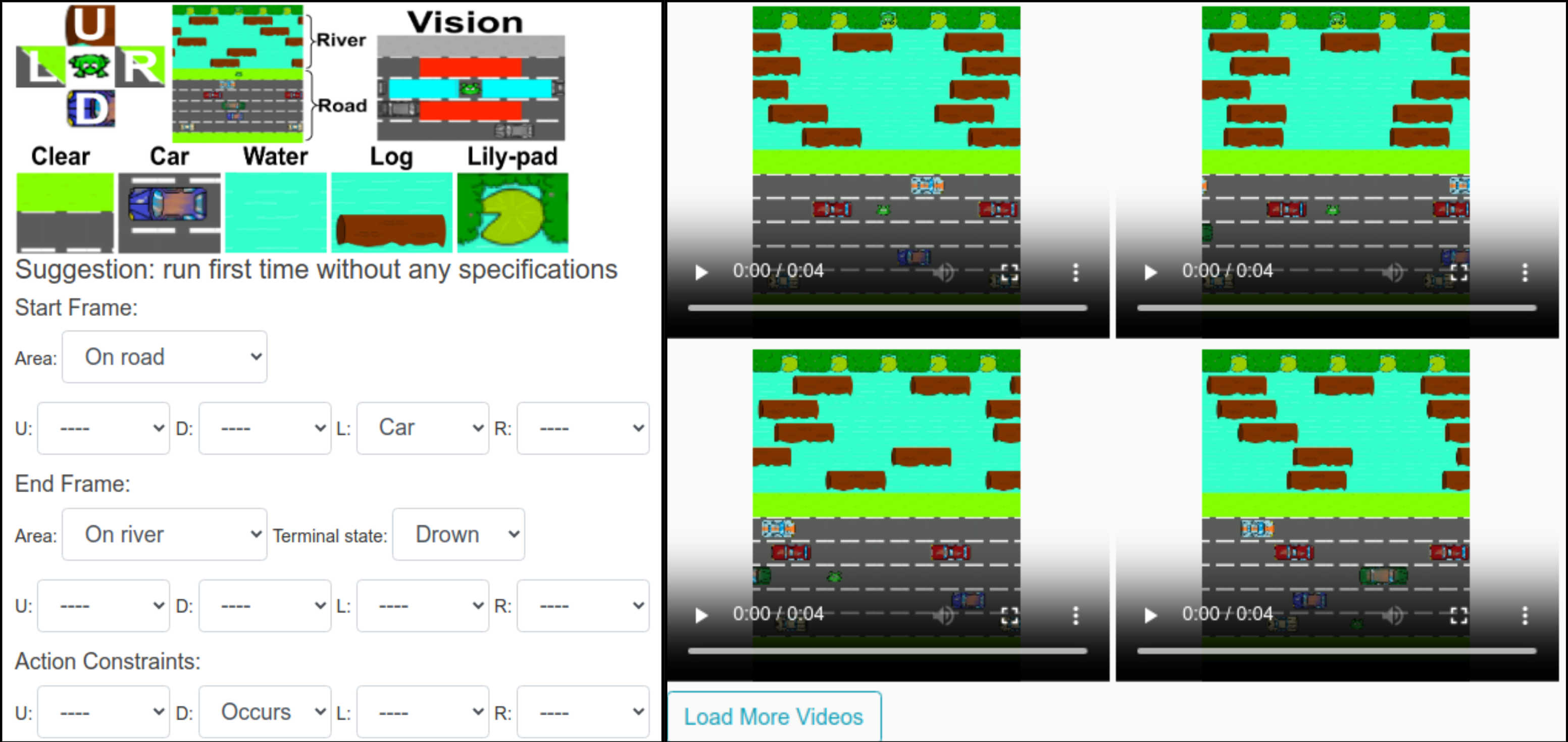}}\\
	\caption{Example ASQ-IT implementation for 
 the Frogger domain. \textbf{Query interpretation:} ``Agent starts on the road with a car on its left and terminates at the river by drowning. Somewhere along its path a DOWN action takes place''}
	\label{fig: frogger interface}
\end{figure} 

\begin{example}{\bf Frogger Domain (Fig. \ref{fig: frogger interface}).}
\normalfont
A user who tries to assess the quality of the agent when crossing the road might query for clips in which the start state is before the road and the end state is before the river. In our experiments, we observed a curious behavior of the agent: it sometimes chooses to move downwards for no apparent reason. A user trying to identify the cause of this behavior might specify the behavior they perceive as odd: query for clips in which at some point, the action ``down'' is taken at a state in which the road immediately ahead of the frog is free. Alternatively, a user might query for clips in which the action ``down'' never occurs. 
\end{example}

\begin{remark}
We stress that the query interface allows users to express a very large number of queries. In our implementation of the query interface for the Highway domain, we allow 5 start positions, 5 start lanes, 6 end positions, 5 end lanes, and 40 different constraint combinations, which amount to \textit{30,000} possible queries. The query interface for the Frogger domain allows \textit{10,935,000,000} different queries. 
\end{remark}

\section{Related Work}
\label{sec:related_work}
This work relates to three main areas of research, which we discuss in this section: (1) explanations in sequential decision-making settings, (2) interactive explanations \& interfaces, and (3) explanations based on formal methods.

\paragraph{Explanations in sequential decision-making settings} In this paper, we focus on the problem of explaining the behavior of agents operating in sequential decision-making settings. Work in this area is typically concerned with explaining policies learned through Reinforcement Learning.

RL explanation methods can be roughly divided into two classes. \emph{Local} explanations focus on explaining specific agent decisions~\cite{krarup2019model,khan2011automatically,hayes2017improving,booth2019evaluating,anderson2020mental}. For instance, ~\citet{greydanus2017visualizing}, visualize the information a game-playing agent attends to in a specific game state by highlighting the saliency of specific pixels on the screen. ~\citet{juozapaitis2019explainable} describe the agent's reward function components in each state through reward-decomposition, meaning how each element of the reward function affects the agent's choice of action in the current state. By understanding the influence and effects of actions on the environment, \citet{madumal2020explainable} construct an action graph able to produce local causal explanations for a given state.

In contrast, \emph{global} explanations aim to convey the agent's MDP, learning scheme, or policy rather than explain particular decisions~\cite{beyret2019dot,shu2017hierarchical,dao2018deep,gottesman2020interpretable,lage2019exploring,huang2019enabling}. 
There are three common approaches to presenting global explanations: (1) Mapping to a more explicit representation, such as generating a proxy model of the policy that is more interpretable, e.g., through policy graphs~\cite{topin2019generation} or decision trees approximating the policy ~\cite{coppens2019distilling}. (2) Extracting rules through logic or structure to create summaries or promote user clarity. For instance, ~\citet{sreedharan2020tldr} summarize, in graphical form, possible agent routes by identifying MDP landmarks, which are defined as environmental conditions, described through propositional formulas, which must be achieved in order for the agent to reach its goal. ~\citet{hein2017particle}, on the other hand, leverage particle swarm optimization on past agent transitions to identify and generate a set of \emph{if-then} rules. (3) Providing demonstrations of the agent's behavior at significant points of interaction to improve users' mental models of how the agent acts~\cite{huang2019enabling}. 
All of these approaches provide static explanations. In this work, we propose an interactive explanation approach drawing on the idea of explanation through demonstrations
of agent behavior. We utilize the idea of extracting demonstrations of agent behavior as a global explanation to answer queries posed by users, such that they can interactively explore the agent's policy and its characteristics. 

\paragraph{Interactive explanations}
Some early work on decision-support systems provided users with interactive explanation methods. For example, MYCIN~\cite{davis1977production}, a system for clinical decision-support, allowed its users to pose ``why'' and ``how'' questions and responded by revealing the rules that led to a particular inference. Such explanations are more difficult to provide in current systems that do not use a logic-based representation. A few approaches in interpretable machine learning have designed interactive explanations for supervised learning models. For instance, TCAV is a method that enables users to test whether the model relies on a user-determined concept in its decision-making~\cite{kim2018interpretability}. Recently, this approach has been applied to analyzing the chess knowledge of AlphaZero~\cite{mcgrath2021acquisition}. 
~\citet{kulesza2011oriented} presented a Why-oriented approach to enable end-users to debug behaviors learned by intelligent assistants through statistical machine learning. The method is primarily focused on Naive Bayes text classification, while ASQ-IT is more broadly applicable to any domain where reinforcement learning agents are used and where understanding their behavior through visual demonstrations is beneficial.
Incorporating insights from ~\citet{verma2020counterfactual} on counterfactual explanations in machine learning, the literature recognizes the role of interactive interfaces. This work, along with others such as ~\citet{granic2017technology} and ~\citet{hohman2019gamut}, emphasizes how interactive designs not only improve user comprehension but also significantly enhance user interaction and engagement with complex algorithms.
Notably, interactive explainable RL has been flagged as a promising research direction in interactive RL research ~\cite{arzate2020survey}.
Most closely related to the problem we discuss, ~\citet{hayes2017improving} leverage statistical analysis to enable users to query an agent regarding when or why an action will take place. A Boolean logic expression that covers most query-relevant states is obtained and converted to text as an explanation. ASQ-IT and Hayes and Shah's (HS) approach both use state predicate-based queries, but HS is confined to a specific query template and provides explicit, text-based explanations from statistical analysis, while ASQ-IT offers a more flexible query space and conveys explanations through visual demonstrations of agent interactions, inviting user interpretation. Additionally, HS's methodology involves learning both a domain model and an agent policy from programmer-specified annotated demonstrations, which require direct access to the agent's code, as opposed to ASQ-IT which assumes no prior knowledge of the agent and treats it as a black box. Moreover, HS employs a graph-search algorithm to fulfill query criteria, with state features selected heuristically, as opposed to ASQ-IT's more user-focused approach of eliciting these from iterative user studies. Furthermore, while HS's approach was demonstrated on example use cases, we evaluate ASQ-IT in two user studies. 
\citet{rupprecht2019finding} and ~\citet{cruz2021interactive} introduce systems to help their users debug agent behavior. Rupprecht's approach involves generating new states and scenarios, potentially unseen or hypothetical, using a generative model to uncover an agent's weaknesses. In contrast, ASQ-IT relies on showcasing videos from existing interactions. Additionally, while ASQ-IT emphasizes engaging users interactively, allowing them to explore and understand the agent's behavior through queries and video responses, Rupprecht's method does not focus on such direct user interaction. Instead, it is geared towards automatically generating these scenarios.~\citet{cruz2021interactive} do utilize an interactive interface, but differ in their approach: In their interface, there is a long video clip of interaction of the agent. The users can pause the video at any timestep, corresponding to a specific world-state in the MDP, and can then query the system about the choice of action in that particular state. In contrast, ASQ-IT focuses on supporting users in finding traces of agent behavior. This cannot be easily done in the interface by Cruz and Igarashi, as a user would need to scan the entire video (which could be very long). The two approaches could potentially be complementary -- ASQ-IT could be used to retrieve traces of interest, and Cruz and Igarashi can then be used to query regarding specific decisions in the trace.

\paragraph{Explanations based on formal methods}
The predominant problem studied in the field of formal verification is, as the name suggests, {\em verification}, namely proving formally that a model of a system satisfies its specification~\cite{handbookMC}. Still, there is active research to develop accompanying tools that aim to explain the system, the specification, or the verification tool's output. In {\em query checking}~\cite{Cha00,BG01,HC22}, the verification engineer inputs a {\em query} $\varphi$, which is a specification  with a ``hole'', denoted as '$?$', and a system $S$ under exploration. The tool finds a formula $\psi$ such that when the hole in $\varphi$ is replaced by $\psi$, then the resulting formula is satisfied by $S$. For example, the query $\varphi = G (? \rightarrow X \ \mbox{fail})$ is read ``it is always the case that $?$ implies that in the next step the system fails''. A formula $\psi$ that can replace $?$ constitutes an explanation for what causes the system to fail. Outputting logic formulas as explanations has also been considered in the context of explaining agent behavior~\cite{kim2014bayesian}.  A conceptually similar approach finds causes for the output of a neural network~\cite{BK23}. {\em Trigger querying}~\cite{KL07} is similar in spirit to query checking only that the tools outputs finite traces of the system as explanations rather than formulas. 
Solving queries is often harder than verification (which is no simple task in its own), and techniques based on heuristics have been developed to allow scalability in practice~\cite{AK14}. Closer to our work, ~\citet{alamdari2020formal} show how to find corner-case traces of an agent that operates in its environment, given a specification from the user and a description of the environment. Their method requires significant manual effort, its scalability is limited, and importantly, it is challenging to adapt to settings like we explore in which only simulation access is given to the agent in its environment. A second line of work explains specifications, which are crafted by engineers and can have mistakes. For example, specifications that are satisfied {\em vacuously} by every system~\cite{BB94,BBER97,KV03}, a specification might not {\em cover} all parts of a system~\cite{CKV06}, and it is important to understand which parts of the system led to satisfaction~\cite{CHK08}. Finally, the counterexample that the verification tool outputs might be hard to understand, and there are tools to assist understanding and repair of the system~\cite{BBCOT12,BD+21,CDF+22}. 
In this work, we draw on formal methods to design an interactive explanation system for RL agents.

\section{The Design of ASQ-IT}
\label{sec:ASQIT_implement}
\begin{figure}[ht]
    \frame{\includegraphics[width=0.9\columnwidth]{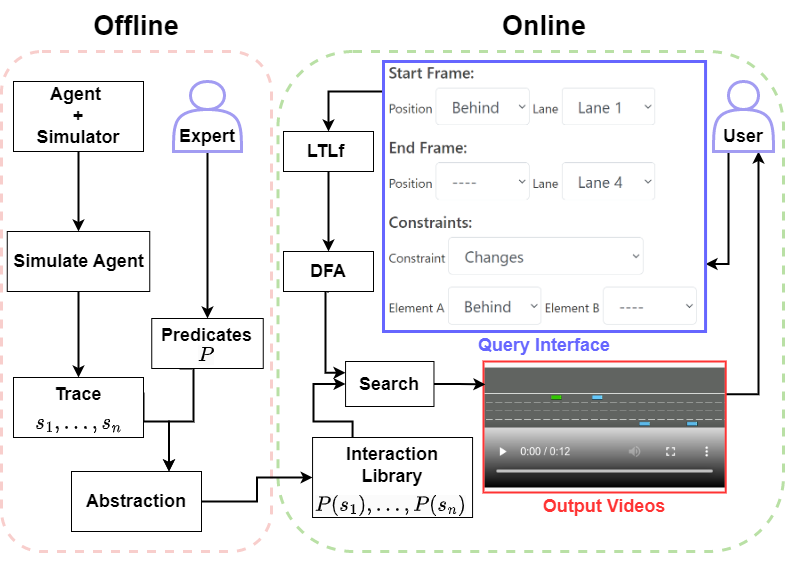}}\\
    \caption{ASQ-IT Process Flow Diagram. \textbf{Offline} - Agent interaction traces are collected from simulations. An abstraction to domain-expert-defined predicates is applied to each state the agent encounters. Abstracted traces are saved to an Interaction Library; \textbf{Online} - User generates query through ASQ-IT's Quert Interface. Query is translated to formal expression and used to search Interaction Library for satisfying sub-traces. Video clips are generated from obtained sub-traces and presented to user.}
	\label{fig: explanation system}
	\vspace{-0.2cm}
\end{figure}

In this section, we describe the building blocks of ASQ-IT\footnote{Code repository at:  \href{https://github.com/yotamitai/ASQ-IT}{link}}. See Figure~\ref{fig: explanation system} for an overview. 
We point out that there are various possible implementations of ASQ-IT, and we highlight the implementation choices throughout this section. These include the domain in which the agent operates, the predicates chosen by the domain expert, and the design of the user interface. 


\subsection{Offline: Obtaining a database of clips}
\label{sec:abstractMDP}
This section discusses the process of obtaining a database of clips for ASQ-IT by simulating an agent in a Markov Decision Process (MDP) environment. It explains how ASQ-IT collects and utilizes a database of state sequences (traces), emphasizing the role of predicates defined by domain experts in interpreting and querying these traces.

ASQ-IT requires simulation access to an agent that operates in an MDP environment. Formally, an MDP is a tuple $\M = \langle S, A, Tr, R \rangle$, where $S$ is a set of states, $A$ is a set of actions, $R: S \rightarrow \Rat$ is a reward function, and $Tr: S\times A \times S \rightarrow [0,1]$ is a probabilistic transition function. An agent acts in the environment according to a {\em policy} $\pi$, which is a function $\pi: S \rightarrow A$. 
The simulator executes $\M$ under the control of $\pi$ and outputs a {\em trace} which is a sequence of states $s_1, s_2, \ldots, s_n$. Note that for $1 \leq i \leq n$, we have $Tr[s_i, \pi(s_i), s_{i+1}] > 0$. We stress that ASQ-IT does not require any knowledge of $Tr$ and $R$, nor of the implementation of $\pi$; all of these are treated as black-boxes. 
In our experiments, $\pi$ is a neural network trained using deep RL. 

We sometimes call a state $s \in S$ of $\M$ a {\em frame} since it can be shown to the user on their screen, and refer to a sequence of states as a {\em clip}. Offline, we collect a {\em database} of traces. For ease of notation, for the remainder of the section, we assume that the database consists of a single trace $s_1,\ldots, s_n$, while in our experiments, we apply the algorithm \ref{alg:backend} on each trace separately. 

\fbox{
\begin{minipage}{0.92\linewidth}
\paragraph{Problem statement (informal)}
Given a database that consists of a trace $s_1,\ldots,s_n$ and a query from the user, output a clip $s_\ell, \ldots, s_k$ that matches the user's query. 
\end{minipage}
}


\paragraph{Domain expert mapping} 
We found, through pilot studies (see Section \ref{sec:queryLang}), that it is infeasible for users to specify a desired behavior directly on the states $S$ of $\M$. Rather, users' queries are formulated on a predefined collection of {\em predicates} $P$ that are chosen by a domain expert. 
Formally, each predicate $p \in P$ is a function $p: s \rightarrow \set{\texttt{True},\texttt{False}}$. ASQ-IT requires an implementation of $p$ that takes a state $s$ and outputs $p(s)$, i.e., whether the predicate $p$ holds at $s$ or not. We denote by $P(s)$, the collection of predicates that hold in $s$, thus $P(s) = \set{p \in P: p(s) = \texttt{True}}$. 
A user's queries will be expressed using $P$. In our front-end design, the drop-down menus allow users to choose between predicates as we elaborate later. 

\begin{example}
\normalfont
We describe the states in the two running-example domains as well as a choice of predicates in these domains.

\textbf{Highway:} 
The agent state within the Highway domain is characterized by a matrix that indicates the speed and position (in a 2D $x,y$ format) of the green car and the $k$ closest cars.\footnote{See `Kinematics` in the highway-env observation documentation: \href{https://highway-env.farama.org/observations/}{link}}

Our pilot experiments revealed that participants often query the agent's absolute position (i.e. the lane it was in) and its relative position compared to other cars (above, behind, in front of, etc.). We chose the predicates accordingly. See the full list of predicates in Fig.~\ref{fig:highway_abstraction}. 

For a state $s \in S$, $\texttt{lane-1}(s) = \texttt{True}$ means that in $s$, the agent is traveling on the top lane, and  $P(s) = \set{\texttt{lane-1}, \texttt{behind}}$ means that at $s$, the agent is traveling in the top lane {\em and} behind some blue car. Conversely, $P(s) = \set{\texttt{lane-2}}$ means that the agent is traveling on Lane~$2$ and there is {\em no} car in front. Otherwise, $\texttt{behind}$ would have been present in $P(s)$. Note that there cannot be a state $s$ for which $P(s) = \set{\texttt{lane-1}, \texttt{lane-2}}$ since the green car cannot be on two lanes at once. 

Recall that queries are formulated using $P$. For example, a query might ask for clips in which the green car travels from Lane~$1$ to Lane~$4$. ASQ-IT's goal would then be to present to the user, a clip $s_k,\ldots,s_\ell$ such that $\texttt{lane-1} \in P(s_k)$ and $\texttt{lane-4} \in P(s_\ell)$.

In order to implement a predicate $\texttt{lane}-l$, for $l=1,2,3,4$, the domain expert determines a range of $y$-coordinates that constitute Lane~$i$ and given $s$, returns $\texttt{True}$ iff the $y$-value of the green car in $s$ belongs to the range. Similarly, the $\texttt{behind}$ predicate is implemented by comparing the $x$-values of the green car with the $x$-values of all other cars whose $y$-value belongs to the same lane. The other predicates are implemented in a similar manner. We note that the domain expert does not actually need to annotate each state. Instead, we implement a function of the defined mapping from raw states to predicates.




\begin{figure}[ht]
    \frame{\includegraphics[width=0.9\columnwidth]{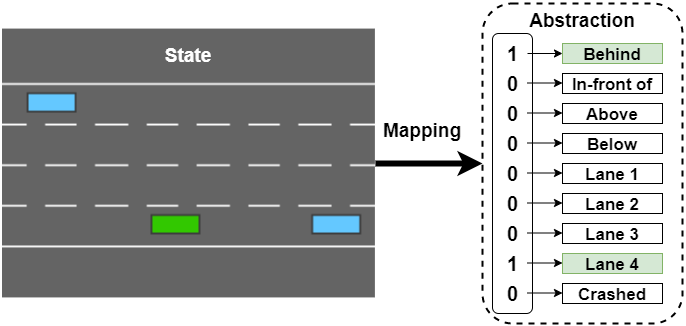}}\\
    \caption{Highway domain state abstraction. \textbf{Left} - An example frame from the simulation. The agent (green rectangle) is in the fourth lane (from the top) and behind another vehicle (blue rectangle); \textbf{Right} - The abstraction of the state to the domain-specific predicates.}
	\label{fig:highway_abstraction}
\end{figure}

\noindent\textbf{Frogger:} 
The state in the Frogger domain includes an array encompassing steps, game level, points, deaths, lives, and the number of frogs on the lilypads. It also details the frog's position (x,y), size (width, height), and orientation, along with similar attributes for each car and log present on the screen. Additionally, the agent has access to the surrounding terrain type, represented by an array where the index indicates direction (e.g., above, below, right...) and the value denotes the terrain type (e.g., road, water, log...).

The predicates that we use in the Frogger domain consist of the location of the frog, e.g., \texttt{before-road}, \texttt{on-road}, \texttt{before-river}, and its surroundings, e.g., \texttt{clear-above}, \texttt{car-below}, \texttt{log-left}. The implementation of these predicates is similar to the Highway domain. In addition, we use the predicates \texttt{action-up}, \texttt{action-down}, \texttt{action-left}, \texttt{action-right}, where for a state $s$, we define $\texttt{action-up}(s) = True$ iff the agent chooses action ``up'' at state $s$, i.e., $\pi(s) = {up}$, and similarly for the other three predicates. This allows users to query, for example, clips in which action ``up'' is never taken, or taken at least once. 
\end{example}


\subsection{Front-end: A formal query language and a user interface}
In this section, we describe the formal basis of our query language. We start by surveying the necessary background on Linear Temporal Logic on Finite Traces (LTLf).We then describe the specific logic used for queries and the design of the user interface.

\subsubsection{Background: Linear temporal logic on finite traces}
\label{sec:LTLf}
Throughout this section, fix a collection of predicates $P$. Recall that $P(s)$ denotes the set of predicates that hold in state $s$. Intuitively, $P(s)$ can be thought of as the ``features'' of state $s$. A query refers to the features of the state and how they develop over time. 
Formally, a user query is an LTLf formula. A formula is satisfied by a sequence $P_1,\ldots, P_m$, where each $P_i \subseteq P$ is a collection of predicates. 
Then, given an LTLf formula $\varphi$ from the user, ASQ-IT searches the database $s_1,\ldots, s_n$ for a clip $s_k,\ldots, s_\ell$ such that the sequence $P(s_k),\ldots, P(s_\ell)$ satisfies $\varphi$.


\begin{example}
\normalfont
We illustrate the syntax and semantics of LTLf. 
Let $P = \set{\texttt{lane-1}, \texttt{behind}}$. 
Consider two sequences 
\begin{itemize}
    \item $\eta_1 = \set{\texttt{lane-1}}, \set{\texttt{lane-1}}, \emptyset$.
    \item $\eta_2 = \set{\texttt{lane-1}}, \set{\texttt{lane-1}}, \set{\texttt{lane-1}, \texttt{behind}}$.
\end{itemize}
Intuitively, both $\eta_1$ and $\eta_2$ describe clips with three frames each. The sequence $\eta_1$ describes a clip in which the agent drives on Lane~$1$ in the first two frames and leaves it (e.g., moves to Lane~$2$) in the third frame. Note that no car drives in front of the agent throughout $\eta_1$. The sequence $\eta_2$ describes a clip in which the agent drives only on Lane~$1$ and in the last frame, a car is in front of it. 

\begin{itemize}
\item The formula $\varphi_1 = F \ \texttt{behind}$ (read ``eventually behind'') specifies clips in which the agent is eventually behind some green car. The sequence $\eta_2$ satisfies $\varphi_1$ but $\eta_1$ does not. 

\item The formula $\varphi_2 = G \ \texttt{lane-1}$ (read ``always Lane~$1$'') specifies clips in which the agent continuously drives on Lane~$1$. The sequence $\eta_1$ satisfies $\varphi_2$ but $\eta_2$ does not. 

\item The formula $\varphi_3 = \texttt{lane-1} \ U \ \texttt{behind}$ (read ``Lane~$1$ until behind'') specifies clips in which the agent drives continuously in Lane~$1$ until it is behind some blue car. The sequence $\eta_2$ satisfies $\varphi_3$ but $\eta_1$ does not. 

\item The formula $\varphi_4 = X \ \texttt{lane-1}$ (read ``next Lane~$1$'') specifies clips in which the agent is driving in Lane~$1$ in the second frame. Both $\eta_1$ and $\eta_2$ satisfy $\varphi_4$. The formula $\varphi_5 = XX \ \texttt{lane-1}$ specifies clips in which the agent is driving in Lane~$1$ in the third frame. The sequence $\eta_2$ satisfies $\varphi_5$ but $\eta_1$ does not. 

\end{itemize}
\end{example}

For completeness, we describe the formal syntax and semantics of LTLf. For more details, see ~\cite{LTLf}. The syntax of LTLf is defined recursively. Each $p \in P$ is an LTLf formula.  If $\varphi_1$ and $\varphi_2$ are LTLf formulas, then so are the following:
\[
\varphi_1 \wedge \varphi_2 \mid \;
\neg \varphi_1 \mid \;
X \varphi_1 \texttt{(read} \: \texttt{`next'} \:\varphi_1 \texttt{)} \mid \;
\varphi_1 U \varphi_2 \: \texttt{(read } \varphi_1 \texttt{ `until' } \varphi_2)
\]

We formally define the operators $F$ and $G$ as abbreviations: 
\begin{itemize}
    \item $F \varphi$ (``eventually $\varphi$'') - is short for $\texttt{True} U \varphi$
    \item $G \varphi$ (``always $\varphi$'') - is short for $\neg F \neg \varphi$.
\end{itemize}


The semantics of LTLf is defined by induction on the structure of the formula. Consider an LTLf formula $\varphi$ over $P$ and an abstract trace $\eta = \sigma_1, \ldots, \sigma_k$, where  $\sigma_i \in 2^P$, for $1 \leq i \leq k$. We say that $\eta$ satisfies $\varphi$, denoted $\eta \models \varphi$, when:

\begin{itemize}
\item If $\varphi =p \in P$, then $\eta \models \varphi$ iff $p \in \sigma_1$.
\item If $\varphi = \varphi_1 \wedge \varphi_1$ then $\eta \models \varphi$ iff $\eta \models \varphi_1$ and $\eta \models \varphi_2$.
\item If $\varphi = \neg \varphi_1$ then $\eta \models \varphi$ iff $\eta \not \models \varphi_1$.
\item If $\varphi = X \varphi_1$ then $\eta \models \varphi$ iff $k >1$ and $(\sigma_2,\ldots,\sigma_k) \models \varphi_1$.
\item If $\varphi = \varphi_1 U \varphi_2$ then $\eta \models \varphi$ iff there is an index $1 \leq i \leq k$ such that $(\sigma_i,\ldots,\sigma_k) \models \varphi_2$ and for each $1 \leq j \leq i$, we have $(\sigma_j,\ldots, \sigma_k) \models \varphi_1$.
\end{itemize}

\paragraph{Nondeterministic finite automata}
A non-deterministic automaton (NFA, for short) is a tuple $\A = \zug{\Sigma, Q, \delta, q_0, Acc}$, where $\Sigma$ is an alphabet, $Q$ is a set of states, $\delta: Q \times \Sigma \rightarrow 2^Q$ is a transition function, $q_0 \in Q$ is an initial state, and $Acc \subseteq Q$ is a set of accepting states. We call $\A$ a {\em deterministic} finite automaton (DFA, for short) when for every $q \in Q$ and $\sigma \in \Sigma$, we have $|\delta(q, \sigma)| \leq 1$. A run of $\A$ on a word $w = \sigma_1 \sigma_2 \ldots \sigma_k$, where $\sigma_j \in \Sigma$, for $1 \leq j \leq k$, is $r = r_0, r_1,\ldots, r_k$, where $r_i \in Q$, for $0 \leq i \leq k$, where $r$ starts in an initial state, i.e., $r_0 = q_0$, and respects the transition function, i.e., for each $i \geq 1$, we have $r_i \in \delta(r_{i-1}, \sigma_i)$. We say that $r$ is {\em accepting} if it ends in an accepting state, i.e., $r_k \in Acc$, and that $\A$ {\em accepts} $w$ if there is an accepting run on $w$. The {\em language} of $\A$, denoted $L(\A)$, is the set of words that it accepts. 

\begin{theorem}
\label{thm:LTLf}
~\cite{LTLf}
Consider an LTLf formula $\varphi$ over a set of predicates $P$. There is a DFA $\A_\varphi$ over the alphabet $\Sigma = 2^P$ whose language is the set of traces that $\varphi$ recognizes. That is, for every trace $\eta \in \Sigma^*$ we have $\eta \in L(\A)$ iff $\eta \models \varphi$. The number of states in $\A_\varphi$ is double-exponential in $|\varphi|$.
\end{theorem}

\subsubsection{A query interface based on a fragment of LTLf}
\label{sec:queryLang}
ASQ-IT is intended for laypeople in logic. 
Attempting to train users in general LTLf would be too steep of a learning curve and would deem the system inaccessible to our target audience. Instead, we follow a standard practice in formal verification called {\em specification patterns} \cite{dwyer1998property,LTLf-patterns,berger2019multiple}. A common challenge in verification is that engineers struggle with using general LTL or LTLf as a specification language. Fortunately, many real-life specifications fall into similar patterns. So, to overcome the specification challenge, an expert identifies common specification patterns. Then, the engineer's task is to choose a pattern and fill in its missing parameters. Both tasks are relatively simple, do not require a deep understanding of the syntax and semantics of general logics, and allow the engineer to specify most real-world scenarios. 

We suggest a specification pattern targeted for querying. To the best of our knowledge, previously identified specification patterns for LTLf are targeted for verification. We developed our pattern based on pilot studies, by analyzing participants' free-text queries and iteratively revising the design of the query interface based on user feedback and analysis. It is meant to be as simple and intuitive as possible yet strong enough to capture many common queries. 


To inform the choice of specification patterns as well as the design of the interface, we conducted several pilot studies and iteratively revised the design of the query interface based on user feedback and analysis.  
Initially, we allowed participants to specify open-ended queries in natural language, prompting them with questions such as ``What types of agent behaviors would you consider to be problematic?'', ``What types of agent behaviors would you consider to be positive?'', and ``What types of agent behaviors would you be curious to see the agent perform?''. This step helped reveal the types of questions users were interested in, however, we found that participants generated queries at ranging levels of granularity and that their queries were often too ambiguous for a formal system to use.

Examples of ambiguous free-form queries provided by participants:
\begin{itemize}
\item ``The Frog maneuvers his way through the road.''
\item ``The Frog takes time before jumping on a log.''
\item ``The Frog gets across the obstacle the fastest.''
\item ``The Frog can hop away from vehicles.''
\item ``The Frog can jump longer''
\end{itemize}

Based on the question types and recurring elements of interest, we determined the desired language for query specifications in formal methods.
We further tested several designs for inputting this information. We began with more free-form designs, such as teaching participants how to define queries in a template-based language and inputting them through text boxes.

Specifically, the template required participants to specify the states of a video of the agent they would be interested in seeing. This required a start and end state but could optionally include intermediate states or constraints between them.
For example, the query ``the frog crosses the river without turning back'' would be roughly written as \texttt{start-state} = ``frog before river'', \texttt{disallowed-actions} = ``down'', and \texttt{end-state} = ``frog above river''. This translates into the LTLf formula: 
\[\texttt{start-state} \rightarrow (\neg \texttt{disallowed-actions}) U \texttt{end-state}\]
With this more structured approach, participants specified more well-defined queries. 
Finally, to constrain the query-space and reduce cognitive load (along with human error of mistyping the queries), a restricted drop-down-based interface was adopted. Additional pilot studies were conducted to refine and optimize ASQ-IT's interactive interface.

The final design of the interface is shown in Figure~\ref{fig: highway interface} (highway domain) and Figure~\ref{fig: frogger interface} (Frogger domain). 
To form a query, a user is required to fill in the following:
\begin{itemize}
\item A description of the start and end state of the trace. These are given as propositional formulas $\varphi_s$ and $\varphi_e$ over the predicates $P$.
\item A constraint on the trace between $\varphi_s$ and $\varphi_e$, which is given as a third propositional formula $\varphi_c$ over $P$.
\end{itemize}

The missing parameters in the pattern are chosen from drop-downs menus. Most of these consist of choosing predicates from $P$ (see Fig.~\ref{fig: highway interface}-Left). Note that a query should be thought of as a constraint; that is, if no descriptions of states are given, ASQ-IT is allowed to output any clip. 


We illustrate how queries are formed in our query interface and describe constraints that we implemented:

\begin{itemize} 
\item Most important is the description of a state by its features (predicates). 
The interface groups predicates into types: one drop-down to select which lane the agent is driving on and a second drop-down to select its relation with other cars (behind, in front, etc). For example, choosing $\texttt{lane-1}$ and $\texttt{behind}$ for the start frame corresponds to setting $\varphi_s = \texttt{lane-1} \wedge \texttt{behind}$, thus ASQ-IT is restricted to output a clip in which in the first frame, the agent is driving on the top lane with a car in front of it. Selecting $\varphi_s = \texttt{lane-1}$ and $\varphi_e = \texttt{lane-4}$ will constraint ASQ-IT to output clips in which in the first frame of a trace, the green car is on the top lane, and in the last frame, the green car is on the bottom lane.

\item The constraint $\varphi_c$ {\em changes} is written in LTLf as $(\varphi_s \wedge \varphi_c) \wedge \ F \neg \varphi_c \wedge \ F G \varphi_e$. For example, given $\varphi_c = \texttt{behind}$ and $\varphi_e = \texttt{lane-4}$, ASQ-IT will output clips that start with the agent driving behind some car, at some point in the clip, the agent is no longer behind a car, and the clip ends with the green car on Lane~$4$.

\item The constraint $\varphi_c$ {\em stays constant} is written in LTLf as $(\varphi_s \wedge \varphi_c) \wedge X (\varphi_c U \varphi_e)$. For example, given  $\varphi_s = \texttt{lane-1}\wedge \texttt{behind}$, $\varphi_e = \texttt{lane-4}$, and $\varphi_c = \texttt{behind}$, ASQ-IT will output clips that start with the agent driving in Lane~$1$ behind some car and ends when the agent is in Lane~$4$, and it drives behind some car throughout the whole clip. 

\item The constraint $\varphi_c$ {\em changes into} $\varphi'_c$ is written in LTLf as $(\varphi_s \wedge \varphi_c \wedge \neg \varphi'_c) \wedge F (\neg \varphi_c \wedge \varphi'_c) \wedge F G\varphi_e$.
For example, given $\varphi_c = \texttt{lane-1}$ and $\varphi'_c = \texttt{lane-2}$, ASQ-IT will output clips that start with the agent driving in Lane~$1$ and at some point switches to Lane~$2$. 
\end{itemize}

\begin{remark}
\normalfont 
As we describe next, our backend is capable of processing general LTLf queries. Thus, it requires minimal effort to enhance the expressivity of the query interface. The only requirement is that each query in the query interface is mapped to an LTLf formula. For example, previous versions of ASQ-IT allowed specifying ``intermediate states'', e.g., specifying a ``zig-zag'' trace in which the green car initially drives on Lane~$1$, visits Lane~$4$, and ends in Lane~$1$.  This property is expressed formally by $\texttt{lane-1} \wedge F (\texttt{lane-4} \wedge F \texttt{lane-1})$. 
\end{remark}


\subsection{Backend: Processing user queries}
\label{sec:algorithm}

We state the algorithmic problem that the backend of ASQ-IT solves: 
Given a query $\varphi$ from the user, our algorithm should efficiently search through the database to find segments that satisfy it. Once a segment is found, we can continue searching the rest of the database, repeating this process until the entire database has been processed.


\fbox{
\begin{minipage}{0.92\linewidth}
\paragraph{Problem statement}
Consider a set of predicates $P$ and a database that consists of a trace $P(s_1),\ldots, P(s_n)$. Given an LTLf query $\varphi$ over $P$, find a sub-trace $P(s_k),\ldots, P(s_\ell)$ that satisfies $\varphi$. 
\end{minipage}
}
\paragraph{The algorithm}
We describe an algorithm (see Algorithm~\ref{alg:backend}) that takes as input a trace $\eta = P(s_1),\ldots, P(s_n)$ over $2^P$ and an LTLf formula $\varphi$, and finds a subtrace $P(s_k),\ldots, P(s_\ell)$ that satisfies $\varphi$. Intuitively, in Lines~$1$-$7$, we search for a position $\ell$ such that the prefix $P(s_1),\ldots,P(s_\ell)$ contains a suffix that satisfies $\varphi$. This is done by constructing an automaton $\A_{F \varphi}$ (read ``eventually $\varphi$'') as in Thm.~\ref{thm:LTLf}. That is, the language of $\A_{F \varphi}$ consists of traces that have a suffix that satisfies $\varphi$. We feed $\eta$ letter by letter to $\A_{F \varphi}$ until it accepts. We point out that when $A_{F \varphi}$ is non-deterministic, we need to maintain a subset of states in which the possible runs of $\A_{F \varphi}$ can be in, and we terminate if one of these runs are accepting. 

Next, in Lines~$8$-$13$, we search for an index $k < \ell$ such that the suffix $P(s_k),\ldots,P(s_\ell)$ satisfies $\varphi$. We ``reverse'' $A_\varphi$ to obtain $\A'_\varphi$: accepting states become initial, initial states become accepting, and transitions are flipped. The language of $\A'_\varphi$ consists of traces that, when mirrored, satisfy $\varphi$. We feed the trace backwards to $\A'_\varphi$ starting from index $\ell$, until the automaton accepts. We point out that $\A'_\varphi$ is guaranteed to accept since the first part of the algorithm ensures that a suffix that satisfies $\varphi$ ends in index $\ell$. Finally, note that $\eta$ is read (forward) once by $\A_{F \varphi}$ and read at most once (backward) by $\A_\varphi$, thus the running time is linear in $n$.


\begin{algorithm}[t]
\caption{Given a trace $\eta = P(s_1),\ldots, P(s_n)$ over $2^P$ and an LTLf formula $\varphi$, find a subtrace $P(s_k),\ldots, P(s_\ell)$ that satisfies $\varphi$}
\label{alg:backend}
\begin{algorithmic}[1]
\State Construct an automaton $\A_{F \varphi}$.
\State $i := 1$
\Do
\State Feed the letter $P(s_i)$ to $\A_{F \varphi}$
\State $i = i+1$
\doWhile{$\A_{F \varphi}$ does not accept the input word}
\State Define $\ell = i$

\State Construct an automaton $\A_\varphi$. Obtain an automaton $\A'_\varphi$ by ``reversing'' $\A_\varphi$.
\Do
\State Feed the letter $P(s_i)$ to $\A'_{\varphi}$
\State $i = i-1$
\doWhile{$\A'_{\varphi}$ does not accept the input word}
\State Define $k = i$
\State \Return $P(s_k),\ldots, P(s_\ell)$

\end{algorithmic}
\end{algorithm}

\begin{theorem}
Consider a collection of predicates $P$ and a trace $\eta$ over $2^P$ of length $n$. Given an LTLf formula $\varphi$, the algorithm returns a sub-trace that satisfies $\varphi$, if one exists. The algorithm processes $\eta$ at most twice.\footnote{We assume that $n$ is much larger than $|\A_\varphi|$ and $|\A_{F \varphi}|$. For short traces, the running time also needs to take into account the size of the queries.} 
\end{theorem}

\begin{remark}
\normalfont 
Once the algorithm finds a trace $P(s_k),\ldots,P(s_\ell)$ that satisfies $\varphi$ it restarts from index $\ell+1$ in search for another query until reaching the end of the database. In our implementation, a query might be answered by numerous clips, dependent on the database size. 
\end{remark}

For user convenience, we defined a lower and upper bound on the length of the traces retrieved. This was used to prune long traces (videos) that fit the query in terms of start and end state but do not actually reflect the user's intention, due to many different behaviors occurring throughout the trace. 
It is interesting to note that in practice, the running time of the algorithm to search for traces is negligible compared to the system's external components such as generating videos\footnote{Video conversion is done using FFmpeg ~\cite{tomar2006converting}} or producing the automata by LTLf to DFA conversion,\footnote{LTLf to DFA conversion is done by MONA ~\cite{monamanual2001} \& LTLf2DFA ~\cite{francescofuggitti_2019}} as seen in Table \ref{tb: runtime}.

\begin{table}[ht]
    \centering
    \small
    \begin{tabular}{c c c}
    \hline
    \textbf{LTLf $\rightarrow$ DFA} & \textbf{Search for Traces} & \textbf{Video Generation} \\
    1.155 $\pm$ .153 & \textbf{0.191} $\pm$ \textbf{.121} & 0.391 $\pm$ .046\\
    \hline
    \end{tabular}
 \caption{Algorithm component runtime in seconds, averaged over 10 queries varying in complexity. Notably, Search runtime is negligible compared to other external components.}
	\label{tb: runtime}
\end{table}

\section{Empirical Evaluation}
\label{sec:evaluation}
To evaluate ASQ-IT, we conducted two user studies. The first study aimed to evaluate ASQ-IT's usability and examine laypeople's ability to understand and use queries in ASQ-IT. The second study aimed to explore how users interact with ASQ-IT in a debugging task. It included both a qualitative analysis of users' interactions with ASQ-IT, as well as a comparison with the use of static policy summaries for the same task. In both studies, we used the Highway domain which is more realistic than the Frogger domain (we used both in pilot studies to ensure our implementation is not limited to a single domain). 
Both studies were approved by the Technion's Institutional Review Board.


\subsection{User study 1: Usability assessment}
\label{sec:usability}
The goal of this study was to examine the usability of ASQ-IT for laypeople, namely, how laypeople comprehend the syntax and semantics of ASQ-IT.

\subsubsection{Empirical methodology}
\emph{Agent.} A double DQN architecture was used to train a Highway domain policy over 2000 episodes, emphasizing safe lane distance, speed range adherence, and collision avoidance.\footnote{The agent implementation is provided in the project code repository.}

\emph{Participants.} Forty participants were recruited through Prolific Academic\footnote{\url{www.prolific.com}, Jul 2022}(20 female, mean age $= 34.7$, STD $= 11.29$), each receiving $\$4.50$ for their completion of the task. To incentivize
participants to make an effort, they were provided a bonus of 15 cents for each
correct answer. The study took $31.6 \pm 10.62$ minutes on average and two participants, whose overall task duration fell more than two standard deviations below the mean, were excluded from the analysis.

\begin{figure}[ht!]
	\centering
    \frame{\includegraphics[width=0.9\columnwidth]{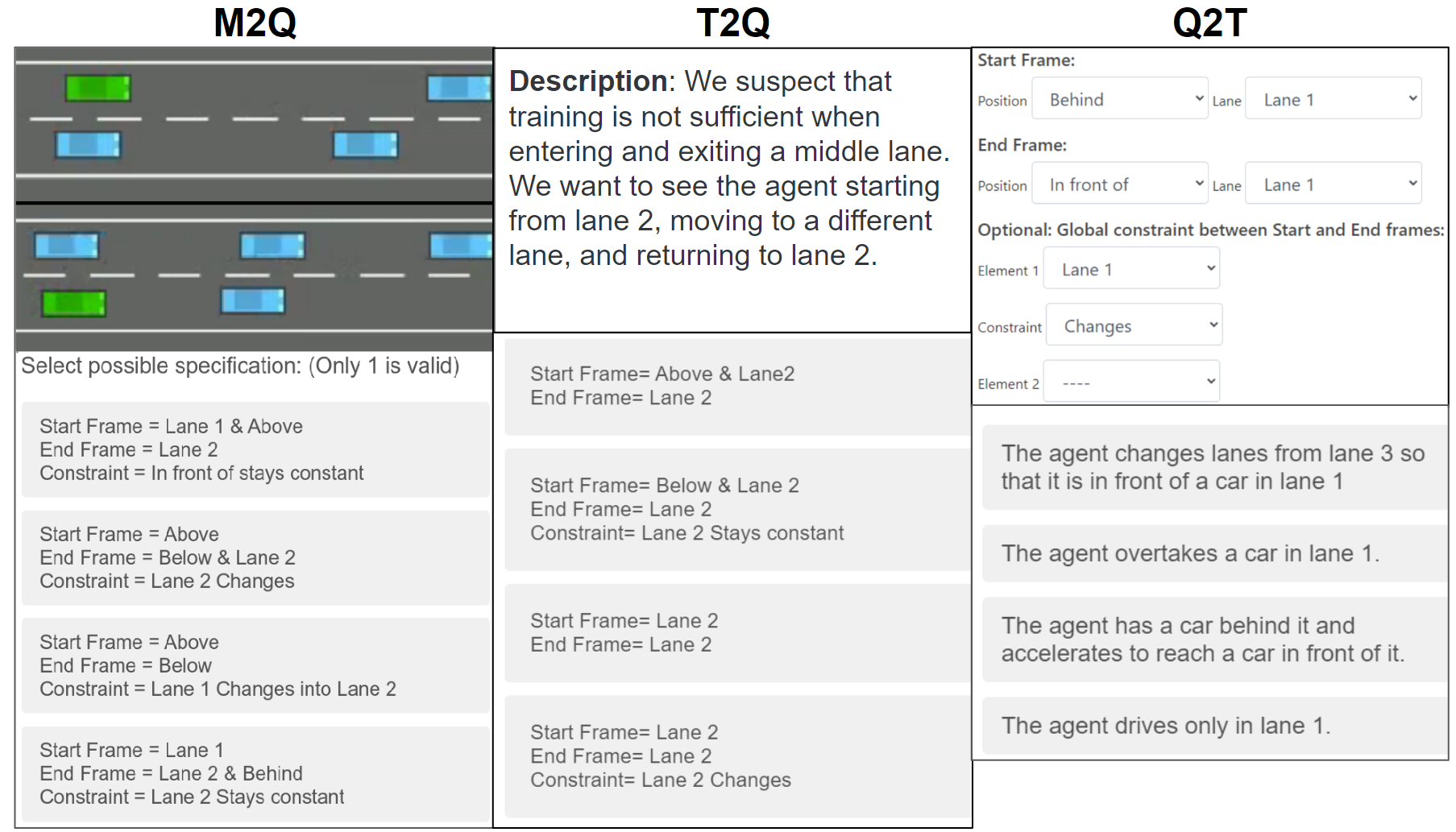}}\\
    \caption{Study 1 example task questions. \\\textbf{Movie to query (M2Q)}: A video clip (depicted here through the first (top) and last (bottom) frames) associated with one of the 4 presented queries. \\ \textbf{Answer}: option 3 (from the top). (Full Video in the \href{https://osf.io/hj3cu/?view_only=9ce523fd224741e3a4767003b146e2b5}{Supplementary Materials}).
    \\ \textbf{Text to query (T2Q)}: A textual description of behavior associated with one of the 4 presented queries. \textbf{Answer}: option 4.
    \\ \textbf{Query to text (Q2T)}: An ASQ-IT query associated with one of the 4 presented textual descriptions of behavior. \textbf{Answer}: option 2.
    }
	\label{fig: M2Q}
\end{figure}

\emph{Tasks.}
We tested participants' comprehension of query syntax and schematics through three types of tasks:
\begin{enumerate}
    \item \emph{Movies to Queries (\textbf{M2Q}):} Given an output video, select the correct query that could result in its generation. 
    \item \emph{Free Text to Queries (\textbf{T2Q}):} Given textual descriptions of desired behavior, select the correct query.
    \item \emph{Queries to Free Text (\textbf{Q2T}):} Given a query, select the correct textual description of the desired behavior.
\end{enumerate}
Examples of all tasks are provided in Fig.~\ref{fig: M2Q}.

\emph{Procedure.}
Before gaining access to ASQ-IT, participants received training that took $\sim$10 minutes. The training process included: (1) an introduction to the Highway domain, (2) an explanation of the concept of AI agents and their possible flaws (e.g., missing training examples, limited training, etc.), (3) a tutorial about ASQ-IT interface, syntax, and semantics, and finally (4) The study tasks.
Each training section was followed by a short quiz to ensure understanding before advancing. Participants were able to return to the instructions to find the answers to the quiz questions and were not able to proceed before answering correctly. As a final step before the task, participants were provided a link to ASQ-IT's interface where they could interact and explore both the interface and the agent until satisfied. 
Then, participants completed the study tasks M2Q, T2Q, Q2T in randomized order. All questions were multiple-choice with four possible answers and a single correct answer. Each task type included two questions in increasing difficulty, quantified through the number and type of elements (e.g.,  queries involving constraints) in the partaking question queries.
Upon task completion, participants were prompted to provide textual feedback regarding their experience with the system and interface and complete a usability survey based on the system usability scale~\cite{brooke1996sus}, by expressing their agreement with statements such as ``I found that the system worked well'' or ``I found the system was easy to use''. The complete user study questionnaire is available in the \href{https://osf.io/hj3cu/?view_only=9ce523fd224741e3a4767003b146e2b5}{Supplementary Materials}.

\emph{Evaluation Metrics and Analyses.}
The study's evaluation involved both quantitative and qualitative analyses. Quantitatively, we measured objective performance based on participants' success rate in M2Q, T2Q, and Q2T questions, along with usability ratings from a 7-point Likert scale survey. These ratings were normalized and averaged, with higher values indicating positive system feedback. Qualitatively, we conducted an analysis of participants' textual responses to identify key insights, common themes and general feedback regarding their experience with ASQ-IT, such as \textit{``Are there questions that you can't ask and want to?''} or \textit{``Did you find certain clips more informative than others when evaluating the agent''}.
Quantitative and qualitative analyses of participants' interaction sessions were performed using recorded sessions and system logs.

\begin{figure}[ht]
	\centering
	\includegraphics[width=1\linewidth]{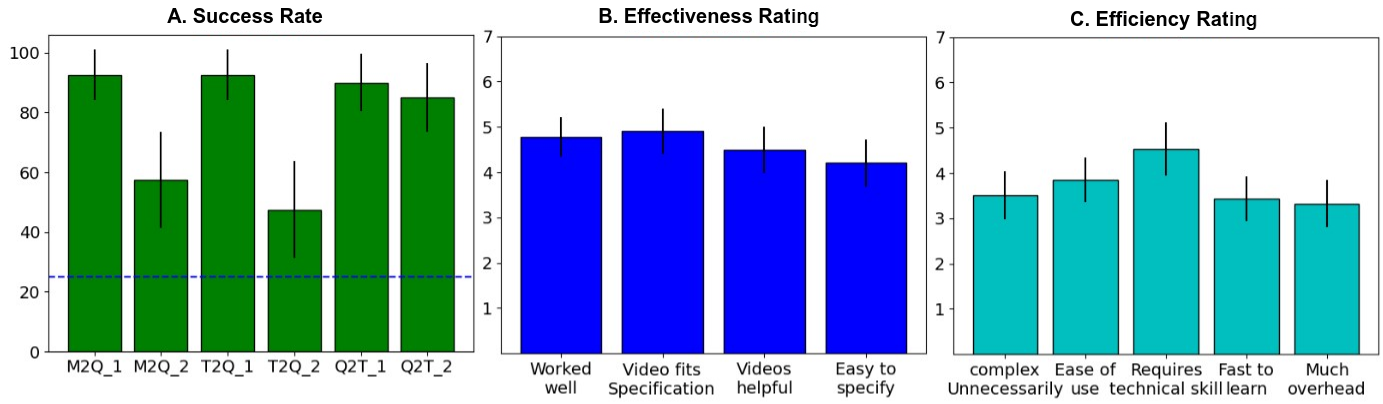}\\
	\caption{User study 1 results: Usability assessment. \textbf{A} - Mean success rate of participant in study tasks; \textbf{B} and \textbf{C} - Mean participant rating of system effectiveness and efficiency (respectively) elicited from post-task usability survey. All graphs include $95\%$ confidence intervals.}
	\label{fig:study1_results}
\end{figure}

\subsubsection{Results \& discussion}
Quantitative results are summarized in Figure~\ref{fig:study1_results}.
We discuss the main findings and insights based on users' responses.

\emph{Participants were able to comprehend the semantics of our logic \& identify relevant queries.}
Overall, participants demonstrated remarkable proficiency in the tasks of interpreting and formulating queries, as shown in Fig.~\ref{fig:study1_results}A. Notably, a one-sample z-test revealed that participants performed significantly better than what would be expected by random chance\footnote{Comparison to random guessing is a common practice in behavioral studies \cite{lord2008statistical}} (choosing out of 4 options, implying a baseline success rate of 0.25, depicted by the dashed blue line in Fig. \ref{fig:study1_results}A). The statistical analysis confirmed that in all questions, participants' success rates exceeded this threshold, with $p < 0.03$. Moreover, in 4 out of 6 questions, the success rate was approximately 90\%. These results are highly encouraging, underscoring the effectiveness of ASQ-IT in enabling participants, who have no training in logic, to express behavior through formal queries in LTLf and to comprehend their output accurately.
 
\emph{Participants improved throughout the task.} 
While many participants described some level of uncertainty upon initial interaction with the interface, mainly citing the lengthy explanations prior to using it, the majority of participants reported quickly understanding once access to ASQ-IT was given and some exploration of the interface was conducted. Some participants who struggled with simple questions regarding constraints managed to correctly answer harder questions that appeared later. Some participants noted that interface elements became clearer when asked to answer questions about them. One participant wrote 
\textit{``I found the instructions quite hard to understand. When a description was provided and you had to complete what you thought was the correct specification, I found this a better way to learn the process.''}
It should be noted that participants had access to the explanation system while completing the task section, such that they could keep exploring and learning about it had they chosen to. 
The explanation system itself operates in a ``Query to Movie (Q2M)'' format, such that it could not have been directly used to answer the questions posed in the task. Overall, participants reported an increase in the ease of using the system over time.

\emph{Some participants displayed difficulties understanding constraints and agent relation syntax.} 
We identified two main causes for incorrect answers: (1) \textit{Agent relations (position):} Confusing the position of the agent compared to other cars such as mixing ``Behind'' with ``In Front Of'' (e.g., is the agent behind another car or is there one behind the agent?), and (2) \textit{Misunderstanding constraints:} Some participants were not able to understand the use of constraints on the agent's trace and most often chose to ignore these specifications.
These alone were responsible for $\approx 90\%$ of all incorrect answers. 
An example can be seen in the M2Q question in Figure~\ref{fig: M2Q}, where all multiple-choice answers have a specification that is plausible given the start and end states, and only the constraint specification dictates the correct answer.

\emph{Participants reported high effectiveness scores.}
In the usability assessment, participants rated ASQ-IT using the system usability scale\cite{brooke1996sus}. The results indicated that ASQ-IT was perceived as significantly more effective than not, in 3 out of 4 categories and marginally significant in the 4th. This was statistically confirmed by conducting a Wilcoxon Signed Rank test for each category, comparing the ratings against a baseline value of $4$. The p-values for each category were $0.003, 0.005, 0.041, 0.59$, respectively, ordered as presented in Fig.~\ref{fig:study1_results}B. These findings underscore the effectiveness of ASQ-IT in terms of usability.

Effectiveness is the measurement of a system's ability to produce the desired outcome. Multiple responses mentioned its usefulness for testing and observing how the agent acts. Others described positively the fact that it was clear to them what videos would be generated by ASQ-IT, so long as the specification was not very complex, and after some initial trial and error phase.
Most negative responses mentioned the many options available and the complexity of understanding the interface. However, many participants reported that after some exploration, their experience and understanding greatly improved, suggesting a learning curve in using the system. One participant described it as the following: \textit{``The system is good but it is difficult to understand at first. It becomes easier the longer you work through and learn it.'' 
}

\emph{Participants reported neutral efficiency scores.} Efficiency is the ability to perform a task successfully, without wasting time or energy. In that sense, system efficiency can be defined as the capacity of a system to perform its designated function in a way that optimizes the use of inputs. Initially, participants experienced uncertainty due to lengthy explanations before using the interface. However, most participants quickly grasped the system after some exploration, indicating a learning curve effect. One participant noted, \textit{``Some of the explanations were a bit confusing but when playing with the system it became easier to understand.''}. When asked for feedback, several participants suggested that the interface could benefit from increased simplicity or having fewer options to interact with, highlighting the trade-off between simplicity and expressivity in design. A Wilcoxon Signed Rank test for each category was conducted to compare to the neutral value of 4. The p-values for each category were 0.08, 0.56, 0.19, 0.04, 0.02 respectively, ordered as presented in Fig.~\ref{fig:study1_results}C with medians between 3 and 4 for all categories, and a median value of 4 aggregated over all scale items.



\emph{Expressivity.}
When asked to describe what features or behaviors were missing or desired for the Highway domain, participants mostly requested the ability to control the agent's speed and distance from other cars, along with the option to specify the positions of other cars and the output video length. 
When asked what agent behaviors and situations were of interest to them, specifiable or not using ASQ-IT's current interface, participants mostly referred to observing the agent react to critical situations such as obstacles on the road, lane merges, or interaction with other cars such as emergency vehicles or evasion of accelerating or braking cars.
These specifications are either complex or impossible to represent in LTLf and we leave these for future work.
Participants also reported their preference for fewer interface options in general, however, there is a trade-off between the simplicity of the interface and its expressivity capabilities.

\subsection{User Study 2: Identifying Agent Faults}
In this study, we sought to assess whether users draw benefits from using ASQ-IT.
The study had two main goals: (1) to understand the process of querying agent behavior using ASQ-IT, and (2) to assess the usefulness of ASQ-IT in a debugging task compared to HIGHLIGHTS, a static policy summary explanation method ~\cite{amir18highlights}.
HIGHLIGHTS was chosen as a baseline for comparison because it offers a global approach that aligns with the study's scope and objectives, and it provides similar outputs in the form of video summaries, ensuring consistency and relevance in the comparative analysis. Additionally, it has been shown to support people's understanding of agent strategies, is computationally efficient, and can be run on the highway domain as opposed to some summary methods that do not scale well to domains with this complexity.
The debugging task was chosen as an example scenario for which we hypothesized that iterative exploration would be beneficial.

\subsubsection{Empirical methodology}
\begin{figure}[ht]
    \centering
    \frame{\includegraphics[width=0.9\columnwidth]{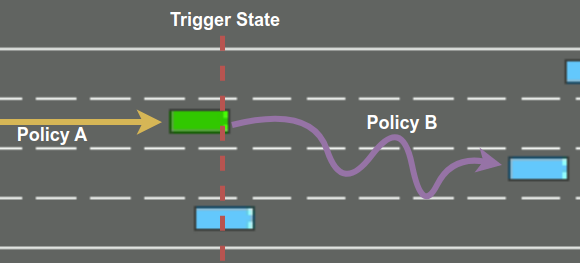}}\\
    \caption{Simulating a faulty agent. The \textit{Plain-Collision} setup:\\ The agent starts with \textbf{policy A} which generates normal driving behavior. Upon reaching a state where predicates \emph{Lane 2} \& \emph{Above} are True, \textbf{policy B} is triggered and takes over, producing an \emph{``out-of-control''} behavior.}
	\label{fig: faulty agents}
	\vspace{-0.2cm}
\end{figure}

The task assigned to participants in this study was to identify a ``trigger'' event that causes a change in the agent's behavior (see Fig.~\ref{fig: faulty agents}). In such a setup we mimic plausible scenarios where agents encounter out-of-training states which can greatly affect their performance~\cite{ramakrishnan2020blind}. For this task, we recruited participants with some basic knowledge of computer science and AI to reduce the initial learning curve of using the system.

\emph{Obtaining agents with triggered dual behavior.} 
To simulate an agent with triggered dual behavior, we initialized concurrently two agent policies $A$ and $B$, with a specified trigger event (e.g., the agent being on Lane~$2$ and above a car). Initially, policy $A$ directs the green car, switching to policy $B$ upon the trigger (Fig.~\ref{fig: faulty agents}). Specifically, we implemented two such dual behavior agents: (1) Used for the elimination task (T1) --  \textit{\underline{Plain-TopLane}}: Policy $A$ is a standard agent (such as was used in Study 1) and policy $B$ prioritizes the top-most lane; and (2) used for tasks 2 and 3 -- \textit{\underline{Plain-Collision}}: Policy $A$ is again the standard agent and policy $B$ attempts to collide with other cars.

All agents were trained for 2000 episodes using the double DQN architecture.

\emph{Participants.} We recruited thirteen graduate students who have completed at least one AI or machine learning course (2 female, mean age $= 29.3$, STD $= 5.1$).
To gain insights into participants' query and thought processes, each participant partook the experiment via a personal interview with the paper's first author, using a think-aloud methodology. The experiment took, on average, 45 minutes to complete. Participants received \$$15$ for their participation. 

\emph{Conditions.} Participants were assigned to either the ASQ-IT explanation system or a system that implemented the HIGHLIGHTS policy summarization algorithm~\cite{amir18highlights}, a method for generating static policy summary explanations. We intentionally assigned more participants for the ASQ-IT condition (8 for ASQ-IT, 5 for HIGHLIGHTS), as we were interested in learning about the interaction with the system.  

\emph{Interface interaction.}
Participants in the ASQ-IT condition interacted with the system through queries constructed using drop-down menus (See Fig.~\ref{fig: highway interface}-Left). Submitting a query would provide participants with up to four videos satisfying the query, chosen randomly from the set of all such videos. An option to load more videos was available given more such videos existed.
Participants in the HIGHLIGHTS condition were presented with a simple interface that only included a single video and an option to load the next video or return to the previous one. Forty videos were made available this way, appearing in a sorted fashion based on the importance assigned to them by the HIGHLIGHTS algorithm. We made use of the HIGHLIGHTS-DIV variant of the algorithm that also takes into consideration the diversity between videos such that the videos were unique and captured multiple important states and not solely the most important ones.
We made sure videos in both conditions videos were of similar parameters such as frame per second (FPS) and minimum length.

\emph{Tasks.}
Participants were tasked with identifying the trigger event that causes the agent to change its behavior. The study consisted of three sub-tasks, as we found that splitting the main task produced a more rigorous manner in participants' performance: 
\begin{enumerate}[(T1)]
    \item \underline{\textit{Elimination}}: Participants explored the \textit{Plain-TopLane} faulty agent using their assigned explanation system. Participants were required to identify the correct trigger from a list of four options and to describe in free text the behavioral change that occurs following the trigger event. 
    \item \underline{\textit{Hypothesis generation}}: Participants were shown two videos of the \textit{Plain-Collision} faulty agent in which the trigger event and the behavior change appear. They were told what the fault was (i.e., trying to collide with other cars) and were asked to propose hypotheses about possible trigger events causing this behavior. Participants were also asked to describe how they would use the system to refute or validate each hypothesis. 
    \item \underline{\textit{Verification}}: Participants were asked to try to refute or validate their proposed hypotheses using the explanation system and, if need be, raise new ones.
\end{enumerate}

\emph{Procedure.}
First, participants were introduced to the Highway domain and its key elements. Second, participants were familiarized with the explanation system they would be using, either the ASQ-IT interface or an interface for watching HIGHLIGHT videos. During this instructions phase, participants could interact with the system and understand how to work it (this was optional, but all participants chose to do so). When satisfied, participants moved on to the study tasks. All tasks included a confidence rating question on a 1 to 7 Likert scale. Lastly, participants answered an explanation satisfaction survey based on ~\citet{hoffman2018metrics}, provided textual feedback on the system they used, and answered demographic questions. The full user study is available in the \href{https://osf.io/hj3cu/?view_only=9ce523fd224741e3a4767003b146e2b5}{Supplementary Materials}. All sessions were done in the presence of the first author who encouraged participants to think aloud. The sessions were recorded, including both the computer screen and the audio. Participants' actions in the system were logged.

\textbf{Example participant interaction:} To illustrate the interaction with ASQ-IT, we detail one participant's (denoted as A3) process throughout the experiment.

\emph{Instructions and 
 Tutorial} (0-9 min) -- Participant A3 was introduced to ASQ-IT and watched a video explaining the system's interface (5 min). Then, was given access to the system interface for self-exploration until satisfied (4 min). 

\emph{T1 - Elimination} (9-22 min) -- After reading the task instructions (2 min), A3 tested the first option \emph{Lane 3 + Below} by setting these predicates in the start state drop-downs. After observing 8 videos without seeing any suspicious behavior, A3 adjusted the query by adding \emph{Lane 1} to the end-state drop-down. When asked ``why?'', A3 responded that they wanted to challenge the agent to see if it could handle the task or possibly observe it fail by crashing. The following clarifications were made: 1) the system will provide videos that satisfy the query, you will not be able to see the agent not being able to execute the query, and 2) a collision was not necessarily defined as the change in behavior.
When option 3, \emph{Lane 4 + Below}, was reached A3 became confident that this was the correct trigger, stating: \emph{``Okay, I think this is the bug -- behind in [Lane] 4 moves to [Lane] 1 always in 12 videos.''}. Due to A3's high confidence, the last option (4) wasn't even tested.

\emph{T2 - Hypothesis Generation} (23-28 min) -- After reading the task instructions and watching the example videos (2 min), A3 was confident the trigger was related to the agent being above another car. They also noted that this occurred in both videos in lane 2 but was not necessarily required, therefore two hypotheses were raised -- 1) \emph{Lane 2+ Above} and 2) \emph{Above}. When asked how they would refute or validate their hypotheses using ASQ-IT, A3 proposed to search both for videos when the agent is above and in lane 2 (to try and reproduce the fault) and ones where it is above but not in lane 2 to try and refute. Additionally, A3 suggested observing random videos to try and better understand what ``normal'' behavior looks like. A confidence score of 5 was selected due to there being only 2 videos.

\emph{T3 - Verification} (29-37 min) -- After reading the task instructions and accessing ASQ-IT's interface (1 min), A3 immediately tested the first hypothesis by defining \emph{Lane 2+ Above} as the start state. The videos observed raised their confidence stating: \emph{``Okay, it looks like it's starting to lose control''}. They then proceeded to query for videos where the agent is above but no lane is defined. Seeing that this provided mostly videos of the agent above in lane 2, A3 tried to further restrict the agent by defining where the agent would start and where it would end (e.g., Start: Lane 3, End: Lane 4). Additionally, A3 also raised a new hypothesis that lane 2 was the trigger and not being above. This too was queried and quickly refuted. 
Finally convinced, A3 provided their final answer that the trigger was their first hypothesis \emph{Lane 2+ Above}, now with an increased confidence of 6.

\emph{Feedback \& Satisfaction} (38-41 min) --
Finally, A3 completed the feedback and satisfaction ratings.

\emph{Evaluation Metrics and Analyses.}
Quantitative and qualitative analyses of participants' interaction sessions were performed using recorded sessions and system logs.

\textbf{Quantitative:}
Participants' success was scored based on their answers' accuracy in identifying the trigger event: (1) No relation - 0 points, (2) Partial relation - 1 point, (3) Exact trigger included among multiple hypotheses - 2 points, (4) Exact trigger chosen - 3 points. In the participants' final answer, where a single hypothesis had to be chosen, scoring 2 points was not possible.

The explanation satisfaction is based on four adapted questions from Hoffman et al.'s survey~\cite{hoffman2018metrics}, tailored to our study's context. These adjustments involved tailoring the survey to the specific task in our study instead of using generic terms like 'goal' or 'task', and omitting items that were not relevant to our research objectives. \cite{amitai2023explaining} made similar adaptations to the survey and show that despite these modifications, the internal consistency of the adapted survey was rigorously maintained as evidenced by high Cronbach's Alpha values, which were greater than 0.9 across all paper studies.

\textbf{Qualitative:} We conducted a qualitative analysis of participants' interactions by coding events in their sessions, including both their interactions with the explanation system and their thought processes.

Initially, the first and second authors independently coded one of the sessions and then discussed their coding to agree on a scheme.  Following discussions among all authors, a coding scheme was established, including categories like 'question', 'answer', 'clarification', 'observation', 'start/stop system interaction', 'hypothesis generation', and 'hypothesis refutal/validation', while noting all significant interactions with the explanation system. The first author applied this scheme to all participant sessions, extracting common themes, observations, and insights.  In addition, the second author coded another two of the sessions independently (altogether three out of the 13 which is $\approx23\%$) and the agreement between them was evaluated (Krippendorff-alpha of $0.93$) We draw on this analysis to identify key themes in participants' interactions with the different explanation systems.

\subsubsection{Results \& discussion}
\begin{figure}[ht]
  \includegraphics[width=0.9\linewidth]{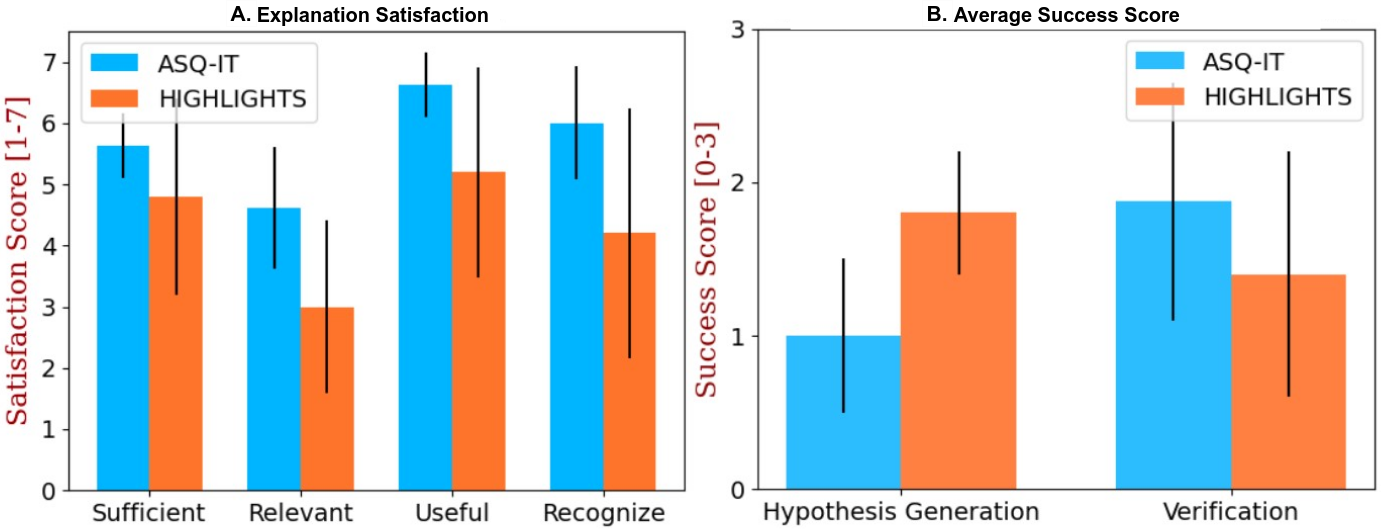}
  \caption{User study 2 results: Agent faults. \textbf{A} - Mean explanation satisfaction ratings; \textbf{B} - Mean change in participant success between the hypothesis generation task (T2) and the verification task (T3).}
  \label{fig:study2results}
  \vspace{-0.4cm}
\end{figure}

Observations from this study indicate a clear advantage of ASQ-IT over HIGHLIGHTS in a debugging task, which manifests both in success score, but more strongly so in participants' experience of actively trying to explore an agent. 
We report the main observations regarding participants' experience and performance with ASQ-IT and compare them to the use of HIGHLIGHTS.
In general, participants using ASQ-IT were more engaged, open to new hypotheses, felt more in control, and were able to explore in a more rigorous manner. In comparison, participants viewing HIGHLIGHTS videos were less active, more frustrated, and less confident. 
We report both the average success scores and the average explanation satisfaction ratings of participants in Figure~\ref{fig:study2results}, as well as a quantization of participant session significant events in Table~\ref{table:study2_analysis}.
Next, we describe the main quantitative and qualitative findings of the study.


\begin{table}[!ht]
    \centering
    \begin{tabular}{|c|c|c|c|c|c|c|c|c|c|c|}
    \hline 
        \multicolumn{2}{|c|}{Phase} & All & \multicolumn{2}{|c|}{Task 2} & \multicolumn{6}{|c|}{Task 3} \\ \hline
         \multicolumn{2}{|c|}{Participant}& Obs & H-G & T2S & H-R & H-V & H-N & Q & V & T3S \\ \hline
        \multirow{9}{*}{\rotatebox{90}{ASQ-IT}} & 1 & 4 & 1 & 1 & 1 & 0 & 1 & 2 & 8 & \textbf{3} \\ 
        & 2 & 5 & 1 & 1 & 0 & 1 & 0 & 3 & 24 & 1 \\ 
        & 3 & 7 & 2 & 2 & 2 & 0 & 1 & 7 & 28 & \textbf{3} \\ 
        & 4 & 9 & 2 & 0 & 2 & 0 & 2 & 7 & 24 & 1 \\ 
        & 5 & 8 & 1 & 1 & 0 & 1 & 2 & 8 & 16 & 1 \\ 
        & 6 & 7 & 1 & 0 & 1 & 0 & 1 & 3 & 20 & 0 \\ 
        & 7 & 29 & 1 & 0 & 1 & 0 & 2 & 9 & 72 & \textbf{3} \\ 
        & 8 & 16 & 1 & \textbf{3} & 0 & 1 & 0 & 12 & 48 & \textbf{3} \\   \hline
        & avg & 10.88 & 1.25 & 1.0 & 0.88 & 0.38  & 1.13  & 6.38 & 30 & 1.88 \\ 
        & stdev &  8.06 &  0.46 & 1.07 &  0.83 &  0.52 &  0.83 &  3.46 &  20.51 & 1.25 \\ \hline
         \multirow{6}{*}{\rotatebox{90}{HIGHLIGHTS}} & 1 & 5 & 2 & \textbf{3} & 1 & 1 & 0 & - & 21 & 1 \\ 
        & 2 & 6 & 2 & 2 & 1 & 1 & 0 & - & 59 & 1 \\ 
        & 3 & 10 & 2 & 2 & 1 & 1 & 0 & - & 30 & \textbf{3} \\ 
        & 4 & 10 & 1 & 1 & 0 & 1 & 0 & - & 18 & 1 \\ 
        & 5 & 27 & 2 & 2 & 2 & 0 & 1 & - & 108 & 1 \\
        & & & & & & & & & & \\ \hline
        & avg  & 11.6  & 1.8 & 2.0 & 1 & 0.8 & 0.2 & - & 47.2 &  1.4\\
        & stdev & 8.91 &  0.45 & 0.45 &  0.71 &  0.45 &  0.45 & - & 37.65 & 0.89 \\ \hline
    \end{tabular}
    \caption{Quantitative analysis of Study 2 participant events, detailing per participant: observations in all tasks (Obs), hypotheses generated in task 2 (H-G), task 2 score (T2S), hypotheses refuted (H-R) and validated (H-V) in task 3, new hypotheses conceived in task 3 (H-N), queries submitted to ASQ-IT in task 3 (Q), videos watched in task 3 (V), and final-answer score in task 3 (T3S).}
    \label{table:study2_analysis}
\end{table}

\emph{ASQ-IT participants revised their original hypotheses.}
Six out of eight ASQ-IT participants revised the hypotheses they generated in the second task based on the output explanation videos produced by their queries to the ASQ-IT interface, while the other two were confident in theirs and chose to keep them.
Meanwhile, only one participant in the HIGHLIGHTS condition revised their original hypothesis. 

\emph{Most ASQ-IT participants who revised their hypotheses improved their identification of the trigger event.}
Out of the six ASQ-IT participants who revised their hypotheses, four were able to improve their score on the final answer. The remaining two participants maintained the same score. In contrast, the HIGHLIGHTS participant who revised her initial hypothesis received a lower score for her final answer compared to her initial response. The average change in participant success is illustrated in Figure~\ref{fig:study2results}B.

\emph{Participants' method of hypothesis verification differed significantly between conditions.}
This was most evident in the \emph{elimination} task (T1).
ASQ-IT participants were able to choose which trigger to inspect, define it as a query, and observe videos of the agent in these situations. They were all able to eliminate options until they reached the correct answer. For six out of eight participants, the correct trigger became immediately evident once queried. The two remaining participants required additional queries to be convinced before ultimately selecting the correct trigger. Apart from one participant, who struggled initially with the interface, mostly due to confusion regarding the role of the constraint drop-downs, all other ASQ-IT participants solved the elimination task quickly and described it as easy.     

HIGHLIGHTS participants, on the other hand, had no control over the videos they received, and as such were forced to see each movie without knowing which trigger option might appear. Four out of five participants' process involved associating each movie with a possible trigger in the list, while the remaining participant searched videos for noticeable patterns and then compared them to the list. Both processes become tedious as the number of options grows, especially when there is no guarantee that any of the trigger options will appear. In referral to their decision process, all participants acknowledged the difficulty of the task, basing their final answers predominantly on the most frequent trigger options observed in the videos, as reflected in the citation (H2):\textit{ ``The elimination wasn’t easy, decision was based on what I saw most''}.

\emph{ASQ-IT participants who identified the correct trigger were able to verify it.}
Out of five ASQ-IT participants who identified the correct trigger (at some point), four were able to verify it using ASQ-IT and submit the correct answer. A typical verification process involved formulating queries that specified hypothesized trigger events and reviewing the retrieved video clips to see whether these indeed led to the behavior change. For instance, participant A7 started way off-course with their original hypothesis being \textit{`` ... whenever the green car falls behind in speed, compared to cars that are in front of it, meaning the cars speed up so it tries to `catch up' to them.''}. However, after multiple interactions, they raised a new hypothesis stating that they \textit{``Noticed the pattern''} and were able to arrive at the correct answer.
Meanwhile, three out of five HIGHLIGHTS participants refuted the correct hypothesis in favor of a more general, but partial answer. This can be associated with the same loss of confidence derived from self-reported lack of control over explanation videos as further discussed below. 

\emph{ASQ-IT participants calibrated their confidence.}
Six out of eight ASQ-IT participants adjusted their reported confidence in a justifiable way based on their interaction with the system. These include two participants who adjusted upwards due to successfully identifying the correct trigger and four participants adjusting downwards based on either the need for revisions or the recognition that the exact answer was not found.
The remaining two participants either reported no confidence change due to recognizing the correct trigger and validating it or were unaware of their partial solution due to confirmation bias which raised their confidence needlessly. 
While interesting, we take this observation with a grain of salt as there are typically substantial individual differences in confidence and the sample size is small.
Four out of five HIGHLIGHTS participants also calibrated their confidence. Three of them lowered their confidence and commented that they were not able to view the videos that they thought would help them validate or refute their hypothesis. One such participant (H5) stated \textit{''Even if I would see 100 videos [of behavior] it wouldn't help me because I need to refute. Watching the same behavior again and again would not help my confidence``}. That is, in contrast to the ASQ-IT participants who lowered their confidence due to observing information that did not align with their hypothesis, HIGHLIGHTS participants lowered their confidence because the system did not provide them with helpful information.

\emph{ASQ-IT participants felt more in control and reported higher satisfaction.}
Upon completion of study tasks, ASQ-IT participants largely reported a positive experience with the explanation system. This positive experience was also evident both in participants' feedback section where they suggested features and options they would like the system to allow in the future (e.g., an ``intermediate state'' to allow querying for more complex behaviors), and orally in their off comments to the experimenter after completing the experiment.
On the other hand, HIGHLIGHTS participants reported more frustration and less satisfaction regarding the explanation system they were assigned, as can be seen in Figure~\ref{fig:study2results}A. All HIGHLIGHTS participants mentioned feeling a lack of control regarding the videos they were shown, four participants stated difficulty in validating or refuting their hypotheses, and three reported loss of confidence. 
One of the participants (H2) concisely summarized these difficulties, stating that \textit{``Lack of variance [in videos] ... hard to refute hypotheses''} and \textit{``Lack of consistency [in videos] ... hard to validate hypotheses''}.

\emph{HIGHLIGHTS participants initially hypothesized the correct trigger more often.}
Four out of five HIGHLIGHTS participants hypothesized the correct trigger solely from observing two videos of the agent, in the \textit{Hypothesis Generation} task (T2), as opposed to only two out of eight ASQ-IT participants. The number of participants does not allow for statistical analysis of this difference, and it is unclear whether it is a result of the differences between the systems.
However, an interesting observation is that while HIGHLIGHTS participants consistently based their hypotheses on examples seen in the first task (elimination (T1)), three out of eight ASQ-IT participants proposed hypotheses relating to behaviors that were never introduced such as speed, agent regret (regarding an action taken) or the act of changing lanes. It might be the case that the systems themselves influence the type of hypotheses generated, with HIGHLIGHTS leading to more direct, example-based hypotheses and ASQ-IT encouraging broader, potentially more complex thinking. This complexity could be advantageous or not, depending on the research goals.

\emph{Participant feedback and improvement suggestions} 
Participant feedback for the ASQ-IT system mainly advocated for added features to enhance expressivity and control, including more predicates, video length customization, and options for intermediate states and multiple constraints. Users suggested improvements like a ``not in'' option for predicates, clearer constraint usage, and functionalities for comparing queries side by side. In contrast, feedback from HIGHLIGHTS users was typically framed as complaints, reflecting a lack of control and frustration from the system. Common points included a need for more context in videos, such as showing events leading up to key moments, as the algorithm tended to highlight states close to crashes without necessarily including the trigger. This disparity in feedback suggests ASQ-IT users are more engaged and offer concrete suggestions, while HIGHLIGHTS users experience a sense of limitation and lack of clarity in system interaction.

\section{Summary and Future Work}
\label{sec:summary_future_work}
With the growing integration of AI systems in sequential decision-making domains, the need for meaningful and engaging explanations is becoming more crucial.
One method for increasing trust while reducing over-reliance on these systems is through interactive interfaces and explanations.
We developed ASQ-IT -- an XRL interactive explanation system for querying AI agents that utilizes formal verification methods. Results from two user studies demonstrate that (1) the system is usable for laypeople with no background in temporal logic, and (2) the system is beneficial for more proficient participants in a debugging task. In the debugging task, the explanation system proved more useful than a baseline static explanation approach, as it enabled users to specify the information that they wished to explore regarding the agent's policy. Beyond the improvement in participants' objective performance in the task, there were noticeable differences in the process of exploring agent behavior. In particular, participants using ASQ-IT were more engaged, open to new hypotheses and felt more in control compared to participants using the static explanation. These findings highlight the potential benefits of designing more interactive explanation methods. We note that there are other tasks for which static explanations could be more useful. For instance, if only a general understanding of the policy is needed, ASQ-IT might add an unneeded cognitive overload.

Some limitations of our experiments should be noted. 
First, the number of participants in Study 2 is relatively small, as each participant required a lengthy think-aloud interview. 
While an experimenter effect was likely present during Study 2, it is crucial to note that it was uniformly applied across all conditions, thereby maintaining comparative validity within the experimental context.
Second, all user studies were conducted on the same domain. While some pilot studies did make use of an additional domain (Frogger), the Highway domain proved extremely useful in designing significantly diverse and non-trivial agent behaviors.
Third, the benefits of our explanation system were only tested on a single task - debugging. Interactivity can benefit users in many different aspects, however, measuring it can prove to be difficult. 
Finally, the benefits of our explanation system were shown for graduate students and not for laypeople. As our system provides a means to explore an agent, it requires some task that would motivate this behavior. This in itself proved non-trivial for participants on platforms such as MTurk or Prolific, who rationally seek to finish tasks quickly and move on. We wanted to elicit observations from the thought process of the explanation system's users and therefore decided to recruit participants with some basic knowledge of logic who could simulate real developers who would seek to debug an agent.

Each of these limitations should be addressed in future work. Additionally, several other directions are worth exploring as well.

A key question in the design of the explanation system is the balance between expressivity and complexity. It is possible that alternative interface designs could provide better scaffolding for more complex queries, such that users could gradually extend their ability to examine policies. Moreover, it would be interesting to go beyond the specifications of state predicates and develop a language for describing more abstract queries about the behavior of the agent (e.g., allowing users to query for ``risky'' behaviors).
Additionally, we plan to explore the incorporation of other temporal logic variants into our framework. This extension would potentially increase the system's versatility in formulating temporal queries and enhance ASQ-IT's ability to process and respond to more complex and abstract user queries.

We aim to do so while maintaining user accessibility, ensuring that these sophisticated elements are presented through an interface that is both intuitive and user-friendly.

An interesting approach that comes to mind given recent advances is the use of large language models (LLMs), such as ChatGPT,\footnote{https://chat.openai.com/chat OpenAI 2023.} to generate the explanation system queries from the user's free-form text. While integrating such tools into ASQ-IT's interface can ease user interaction, it might reduce overall performance by not bounding users' queries as the drop-downs do. Recent work ~\cite{buccinca2021trust} has also demonstrated the benefits of cognitive forcing functions in reducing over-reliance. Keeping these matters in mind, integrating LLMs is a natural future direction we intend to pursue.

The ASQ-IT system could also benefit from incorporating a range of existing explanation methods, allowing users to switch between various predefined explanations like HIGHLIGHTS and formulating their own queries. Moreover, we envision that users start with HIGHLIGHTS to obtain some insight regarding an agent's global behavior and aptitude, and then progress to ASQ-IT for further detail and deeper exploration of specific behaviors. This integration would not only diversify the explanatory tools available but also assist users in pinpointing specific aspects of the agent's policy that merit deeper investigation. By providing users with a blend of pre-specified explanations and the flexibility to explore custom queries, the system can enhance user understanding and engagement with the agent's decision-making processes. Moreover, ASQ-IT can also integrate with local explanation methods (e.g., reward decomposition, saliency maps) which can add information regarding the decisions made by the agent in the shown trajectories.

\section{Acknowledgments}
This research was partially funded by the Israeli Science Foundation grant \#2185/20 and \#1679/21, ERC Starting grant \texttt{\#101078158} CONVEY.
Views and opinions expressed are however those of the author(s) only and do not necessarily reflect those of the European Union or the European Research Council Executive Agency. Neither the European Union nor the granting authority can be held responsible for them.
OA also acknowledges the support of the Schmidt Career Advancement Chair in AI.

\bibliographystyle{plainnat}  
\bibliography{main}

\end{document}